\documentclass[11pt]{article}

\usepackage[T1]{fontenc}
\usepackage[utf8]{inputenc}
\usepackage{times}
\usepackage[margin=1in]{geometry}
\usepackage{microtype}

\usepackage{amsmath,amssymb,amsthm}

\usepackage{amsmath,amsfonts,bm}

\def\eqref#1{equation~\ref{#1}}

\def\1{\bm{1}}

\def\vh{{\bm{h}}}

\def\vq{{\bm{q}}}

\def\vy{{\bm{y}}}
\def\vz{{\bm{z}}}

\DeclareMathAlphabet{\mathsfit}{\encodingdefault}{\sfdefault}{m}{sl}
\SetMathAlphabet{\mathsfit}{bold}{\encodingdefault}{\sfdefault}{bx}{n}

\usepackage{graphicx}
\usepackage{booktabs}
\usepackage{array}
\usepackage{multirow}
\usepackage{tabularx}
\usepackage[table]{xcolor}
\usepackage{capt-of}
\usepackage[most]{tcolorbox}
\usepackage{svg}
\usepackage{fvextra}

\usepackage{algorithm}
\usepackage{algpseudocode}

\usepackage[round,authoryear]{natbib}
\usepackage{url}
\usepackage[
    colorlinks=true,
    linkcolor=blue,
    citecolor=blue,
    urlcolor=blue
]{hyperref}

\newcolumntype{C}{>{\centering\arraybackslash}X}

\title{
    Know Thyself, Teach Thyself: Internal Information Flow for Selective Self-Distillation
}

\author{
    Rui Wang\textsuperscript{1}\thanks{
        \texttt{2300010607@stu.pku.edu.cn}
    }
    \quad
    Ruijie Wang\textsuperscript{2}
    \quad
    Bo Chen\textsuperscript{3}
    \quad
    Jiangxuan Long\textsuperscript{3}
    \quad
    Yingyu Liang\textsuperscript{3}\thanks{
        \texttt{yingyul@hku.hk}
    }
    \\[0.7em]
    \textsuperscript{1}Peking University
    \qquad
    \textsuperscript{2}University of Oxford
    \qquad
    \textsuperscript{3}The University of Hong Kong
}

\date{}

\newcommand{\method}{\textsc{InFlow}}

\newcommand{\KLdiv}{\operatorname{KL}}

\newtheorem{proposition}{Proposition}

\definecolor{stdgray}{gray}{0.94}

\begin{document}

\maketitle

\begin{abstract}
Self-distillation turns knowledge distillation into a closed learning loop and offers a path toward recursive self-improvement. Without an external teacher, however, the model must determine both what information can improve its supervision and which induced changes should be learned. Existing methods typically improve teacher-generated data or select training examples in isolation, leaving the information transferred between these stages unmeasured. We introduce \method{}, a retrieval-guided on-policy self-distillation framework that models this process as potential-to-realized information flow. \method{} first retrieves potentially informative sources using certainty-calibrated hidden-state trajectories, then measures their realized effect through the Jensen--Shannon divergence between the teacher's initial and retrieval-conditioned answer beliefs. Examples with larger belief shifts are selected for on-policy distillation. Our analysis formalizes the information optimized by retrieval and selection and relates the answer-level shift to the teacher--student distillation gap. Across four open-weight language models and three knowledge domains, \method{} achieves the strongest cross-model average among the compared selection methods, with ablations supporting both stages of the framework. Our code is available at \url{https://github.com/1240148048/INFLOW}.
\end{abstract}

\section{Introduction}

Knowledge distillation transfers predictive information from a teacher to a student, including relations among outputs and representations that hard labels discard~\citep{hinton2015distilling,ojha2023knowledge,zhang2023information}. Self-distillation retains this principle while replacing the external teacher with another view of the learner, obtained from a previous checkpoint, self-generated data, or richer conditioning~\citep{tarvainen2017mean,furlanello2018born,zhang2019byot}. This removes the dependence on a larger teacher, makes unlabeled data directly usable, and keeps supervision within the learner's own output space. For language models, self-generated instructions and reasoning traces have extended self-distillation from regularization to iterative adaptation~\citep{wang2023selfinstruct,zelikman2022star,huang2023selfimprove,chen2024self,yuan2024selfrewarding}. Such a closed loop offers a path toward \emph{Recursive Self-Improvement (RSI)}: the prospect that a model can repeatedly transform its own experience and available evidence into supervision for a more capable successor, without requiring a progressively stronger external teacher or human supervision.

The same autonomy creates the central obstacle. Without an independent teacher or gold labels, self-generated supervision inherits the model's uncertainty and systematic errors, which subsequent optimization may reinforce. Existing methods approach this problem from two directions. Teacher-data enhancement constructs a stronger model view through retrieval, contextual guidance, self-consistency, revision, or self-reward~\citep{wang2023selfconsistency,yuan2024selfrewarding,ye2026policy,hubotter2026reinforcement}. Data-selection methods instead retain examples according to confidence, diversity, influence, consistency, or internal activation~\citep{liu2024deita,xia2024less,wang2023scott,chen2026neuron}. Annotation-free methods perform these operations without labeled targets, while on-policy distillation aligns teacher guidance with prefixes visited by the current student~\citep{huang2023selfimprove,agarwal2024onpolicy}. These advances improve either the information supplied to the teacher or the supervision consumed by the student, but usually optimize the two decisions separately.

This separation overlooks that distillation is fundamentally a process of information transfer~\citep{ojha2023knowledge,zhang2023information}. Progress toward RSI requires a model to diagnose its current decision state from internal representations, retrieve information targeted to that state, ruminate on the augmented teacher-generated data, then filter it, and train on the effective information. This creates a directed flow in which external evidence enters the contextual teacher through retrieval, and the information expressed by the revised teacher returns to the student through selective distillation. To our knowledge, existing self-distillation methods have not connected \textbf{model-internal representations} with \textbf{end-to-end optimization of this information flow}. Conditional mutual-information objectives provide a related perspective in knowledge distillation~\citep{ye2024bayes,chen2024mutual}, but they rely on teacher-side information or external verification that do not capture the endogenous, intervention-dependent flow of annotation-free self-distillation.

We introduce \method{} (\textbf{In}ternal Information \textbf{Flow}) following the principle of maximizing the information expected to be transferred during self-distillation. Before generation, certainty-calibrated hidden-state trajectories identify sources that can provide the largest potential information to the target's current decision state. After retrieval, \method{} compares the teacher's initial and contextual answer beliefs, using their Jensen--Shannon divergence to measure the semantic information realized by the intervention. Examples with larger realized shifts are retained for on-policy distillation, which transfers the retrieval-conditioned teacher distribution back to the student. The resulting potential-to-realized-to-distilled pipeline connects targeted information acquisition, teacher self-comparison, and selective model updating within one optimization framework. Figure~\ref{fig:pipeline} presents an overview of the entire INFLOW pipeline. Our contributions are:

\begin{figure}[t]
    \centering
    \includegraphics[width=\linewidth]{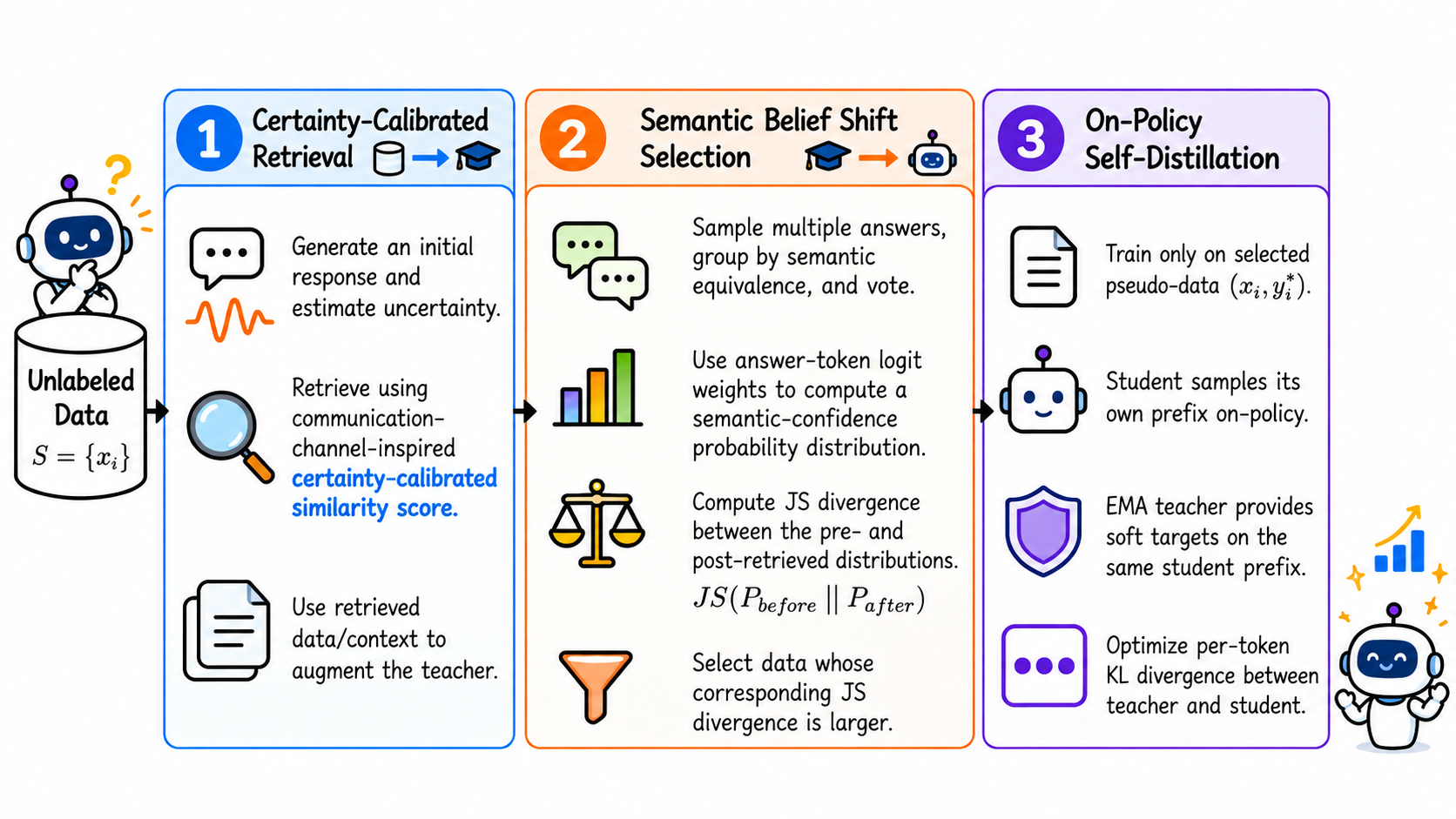}
    \caption{Overview of \method{}. For each query, Stage~I uses internal signals to estimate how much information retrieval can channel from external evidence into the teacher. In Stage~II, the teacher reflects through self-comparison and selects the realized information to provide to the student. Stage~III distills this new information into the student, where it is expected to shift the original semantic belief without retrieval context.}
    \label{fig:pipeline}
\end{figure}

\begin{itemize}
    \item \textbf{Method.} We formulate annotation-free self-distillation as a closed information-flow problem and introduce a coupled mechanism that retrieves potentially informative internal sources, measures their realized effect on the teacher, and distills the selected contextual supervision.

    \item \textbf{Experiments.} Across four model backbones and three knowledge domains, \method{} improves the corresponding unadapted checkpoints by $1.0$--$3.2$ overall accuracy points and achieves the strongest cross-model average among the compared selection methods. Controlled ablations further analyze the contribution of each module. Comparative experiments demonstrate the generalization of our method across different hyperparameter settings.

    \item \textbf{Theory.} We show that \method{} jointly uses certainty-calibrated retrieval and semantic belief shift to optimize the budgeted allocation of information flow for self-distillation given the available evidence. We further characterize the convergence rate of the teacher--student information gap during distillation and qualitatively discuss the ability and limitations of semantic belief shifts to predict correct-label revisions without ground-truth supervision.
\end{itemize}
\section{Related Work}

\subsection{Self-Distillation and Iterative Self-Improvement}

Self-distillation extends knowledge distillation by replacing the external teacher with alternative views of the model itself, such as averaged checkpoints, successive generations, or intermediate representations~\citep{tarvainen2017mean,furlanello2018born,zhang2019byot}. Language models further externalize these views as generated instructions, shifting self-distillation from representation regularization toward iterative adaptation~\citep{wang2023selfinstruct,zelikman2022star,huang2023selfimprove}. Annotation-free methods remove the need for labeled targets, while on-policy methods align supervision with states visited by the current student~\citep{chen2024self,yuan2024selfrewarding,agarwal2024onpolicy}. Together, they suggest a path toward continual self-improvement, in which each updated model produces supervision for the next round.

The central obstacle is that this supervision remains endogenous. Self-training can propagate confidently incorrect pseudo-labels across iterations~\citep{rodemann2023bayes}; without external feedback, recursive training on generated data can accumulate approximation errors and progressively discard low-probability regions of the original distribution~\citep{huang2024selfcorrect}. More broadly, repeated training on data generated for a particular task, or poorly designed difficulty levels, may lead to overfitting and catastrophic forgetting~\citep{shumailov2024collapse}. Sustainable self-improvement therefore places requirements on both data quality and curriculum design.

\subsection{Teacher Data Enhancement and Selection for Self-Distillation}

One line of work improves the supervision produced by the teacher before it reaches the student. Self-Instruct expands the training distribution through model-generated instructions, while STaR iteratively retains successful reasoning traces to strengthen subsequent supervision~\citep{wang2023selfinstruct,zelikman2022star}. Later approaches improve teacher outputs through agreement across generations, contrastive rationales, persistent behavioral consistency, or self-evaluation and revision~\citep{wang2023scott,yuan2024selfrewarding,lv2026pcsd}. Retrieval offers another route: methods such as KATE and EPR construct a more informative teacher context by supplying demonstrations that are semantically relevant or preferred by the language model~\citep{liu2022kate,rubin2022epr}. These mechanisms are particularly attractive for self-distillation because they can create a stronger \emph{teacher view} without introducing a separate, more capable model. Their reliability, however, is limited by the closed nature of self-supervision. Semantic proximity does not imply that a demonstration contains a reliable decision process, and self-evaluation may inherit the generator's own blind spots. Therefore, improving teacher data quality under unsupervised conditions requires not only enhanced information acquisition and internal self-evaluation, but also prediction of its actual effectiveness during training.

A complementary line of work selects which teacher-generated examples should drive optimization. DEITA balances estimated data quality against redundancy, LESS measures how a candidate's gradient influences a target task, and NEURON uses model-internal activation patterns to select examples without annotations~\citep{liu2024deita,xia2024less,chen2026neuron}. Selection principles developed for conventional knowledge distillation can also be transferred to self-distillation. For example, consistency-based rationale transfer and mutual-information objectives motivate ranking a candidate demonstration by how much information it provides about the target answer $Y$ beyond the input~\citep{wang2023scott,chen2024mutual}. This transfer is not direct, however. Conventional distillation usually relies on a fixed and independently trained teacher, so teacher--student agreement or dependence has a meaningful external reference; in self-distillation, the teacher and student share parameters, architecture, and often the same errors. Consequently, high consistency or mutual information may indicate redundant copying. Therefore, during the data selection phase of self-distillation, ``dissenting voices'' may be more valuable than highly similar information.
\section{Method}
\label{sec: method}

\subsection{Problem Setting and Overview}

Let $\mathcal{U}=\{x_i\}_{i=1}^{N}$ be an unlabeled collection of questions and let $p_{\theta}$ be a causal language model. Each question contains a prompt and, for multiple-choice tasks, a legal answer set $\mathcal{A}_i$. We seek an adapted model without gold labels or a stronger external teacher. One iteration of \method{} contains three stages: (1) certainty-calibrated retrieval, (2) semantic belief-shift selection, and (3) on-policy distillation from an EMA teacher. Retrieval and selection are recomputed from the model's own signals without external annotations.  Algorithm~\ref{alg:inflow} in Appendix~\ref{app:algorithm} gives the complete procedure.

For each $x_i$, the model first emits a structured response containing a short knowledge summary, a solution method, the selected option text, and its answer label. We then run a forward pass over the prompt and this initial response. Let $t_i$ be the position of the final answer token and let
\begin{equation}
    \tau_i = \left(\bar{\vh}_{i,t_i}^{(1)},\ldots,\bar{\vh}_{i,t_i}^{(L)}\right),
    \qquad
    \bar{\vh}_{i,t_i}^{(\ell)}=
    \frac{\vh_{i,t_i}^{(\ell)}}{\|\vh_{i,t_i}^{(\ell)}\|_2},
    \label{eq:trajectory}
\end{equation}
where $\vh_{i,t_i}^{(\ell)}$ is the hidden vector at the same final-token position after transformer layer $\ell$. Thus $\tau_i$ is the trajectory followed by the decision token as information is transformed through depth.

\subsection{Certainty-Calibrated Retrieval as Potential Information Flow}

We flatten the layer axis only when computing a pairwise cosine:
\begin{equation}
    s_{ij}=\max\!\left(0,
    \frac{\langle \operatorname{vec}(\tau_i),\operatorname{vec}(\tau_j)\rangle}
    {\|\operatorname{vec}(\tau_i)\|_2\|\operatorname{vec}(\tau_j)\|_2}
    \right).
    \label{eq:similarity}
\end{equation}
The nonnegative part excludes trajectories whose layerwise directions are opposed. 

Similarity identifies aligned computation but does not establish that the source decision is trustworthy. Let $\vq_j$ be the softmax distribution over the legal answer-token logits of source $j$. We use Shannon entropy to quantify uncertainty, normalized to a score between 0 and 1. The certainty score is then defined as one minus the uncertainty score. The certainty score is zero for a uniform answer belief and approaches one when probability mass concentrates on one option. 
\begin{equation}
    c_j = 1-\frac{H(\vq_j)}{\log |\mathcal{A}_j|},
    \qquad
    H(\vq_j)=-\sum_{a\in\mathcal{A}_j}q_j(a)\log q_j(a).
    \label{eq:certainty}
\end{equation}

Channel theory suggests that a useful demonstration should provide a strong signal about the target decision through a reliable channel. To derive the retrieval score, let $T_i\sim\mathcal{N}(0,1)$ denote a local scalar representation of the target decision and source $j$ as the standardized Gaussian observation
\begin{equation}
Z_{ij}
=
\sqrt{c_j}\,s_{ij}T_i
+
\sqrt{1-c_js_{ij}^{2}}\,\varepsilon_j,
\qquad
\varepsilon_j\sim\mathcal{N}(0,1).
\label{eq:local-channel}
\end{equation}
The effective squared correlation is therefore
$\rho_{ij}^{2}=c_js_{ij}^{2}$: certainty controls the probability of reliable transmission, while squared similarity determines its aligned signal strength. The theoretical information of this local Gaussian channel is
\begin{equation}
I_{ij}^{\mathrm{pot}}
=
I(T_i;Z_{ij})
=
-\frac{1}{2}\log\left(1-\widetilde{\rho}_{ij}^{\,2}\right),
\qquad
\widetilde{\rho}_{ij}^{\,2}
=
\min\{\rho_{ij}^{2},1-\epsilon\}.
\label{eq:potential-information}
\end{equation}
We use this quantity as a local information-potential surrogate and retrieve the top $k$ sources for each target. Because $I_{ij}^{\mathrm{pot}}$ is monotone in $c_js_{ij}^{2}$, this ranking favors sources whose internal trajectories are both strongly aligned with the target and supported by a reliable answer belief. The score is directional because reversing $i$ and $j$ generally changes the estimated channel quality.

\subsection{Semantic Beliefs and Realized Belief Shift}
\label{sec:belief-shift}

For each target $x_i$, we compare its initial decision with the teacher responses induced by the retrieved set $S_i$. We reuse the single no-retrieval response generated in Stage I, and generate $M$ retrieval-conditioned responses using the same decoding configuration,
\begin{equation}
y_i^0 \sim p_{\bar{\theta}}(\cdot\mid x_i), \quad
y_i^{S,m}
\sim
p_{\bar{\theta}}(\cdot\mid x_i,S_i),
\qquad
m=1,\ldots,M.
\end{equation}
Let $t_i^0$ and $t_i^{S,m}$ denote the final answer-token positions of these responses. For each legal answer $a\in\mathcal{A}_i$, let $z_i^0(a)$ and $z_i^{S,m}(a)$ be the corresponding teacher logits. We normalize these logits:
\begin{equation}
q_i^0(a)
=
\frac{
\exp\!\left(z_i^0(a)/T\right)
}{
\sum_{a'\in\mathcal{A}_i}
\exp\!\left(z_i^0(a')/T\right)
},
\qquad
q_i^{S,m}(a)
=
\frac{
\exp\!\left(z_i^{S,m}(a)/T\right)
}{
\sum_{a'\in\mathcal{A}_i}
\exp\!\left(z_i^{S,m}(a')/T\right)
},
\label{eq:legal-answer-belief}
\end{equation}
where $T$ is the default decoding temperature. Normalization over the legal answer set makes each distribution invariant to an additive shift of the vocabulary logits at its scoring position. For multiple-choice tasks, the initial semantic belief is the answer-token distribution from the single no-retrieval response, while the retrieval-conditioned belief averages the distributions of the $M$ teacher responses:
\begin{equation}
P_i^0(a)
=
q_i^{0}(a),
\qquad
P_i^S(a)
=
\frac{1}{M}
\sum_{m=1}^{M}
q_i^{S,m}(a).
\label{eq:semantic-belief}
\end{equation}
For open-ended generation tasks, where no fixed legal answer set is available, we follow the semantic-entropy principle of grouping semantically equivalent responses and estimating probability mass in the resulting semantic space~\citep{kuhn2023semantic}; Appendix~\ref{app:free-form-belief} provides the corresponding estimator.

This construction compares the model's initial sampled decision with the distribution of teacher decisions induced by the retrieved context. The realized belief shift is
\begin{equation}
B_i
=
\operatorname{JS}(P_i^0\Vert P_i^S)
=
\frac{1}{2}\operatorname{KL}(P_i^0\Vert M_i)
+
\frac{1}{2}\operatorname{KL}(P_i^S\Vert M_i),
\qquad
M_i=\frac{P_i^0+P_i^S}{2}.
\label{eq:belief-shift}
\end{equation}
INFLOW retains the top $\alpha N$ targets according to $B_i$. The score measures the magnitude of the retrieval-induced change in the teacher's semantic belief, and serves as an estimate of the information transferred to the student through distillation on that example. Sec.~\ref{sec: discussion} shows that INFLOW follows the principle of maximizing the expected amount of information transferred through distillation.

\subsection{On-Policy Self-Distillation}

For a selected target, the student samples a trajectory $\hat{\vy}\sim p_\theta(\cdot\mid x_i)$. At every student prefix $\hat{\vy}_{<t}$, the EMA teacher is conditioned on the same prefix and retrieved demonstrations $S_i$. The update minimizes reverse KL divergence:
\begin{equation}
    \mathcal{L}_{\mathrm{OPSD}}=\frac{1}{|\hat{y}|}\mathbb{E}_{\hat{\vy}\sim p_\theta}\sum_{t=1}^{|\hat{\vy}|}\KLdiv\!\left(p_\theta(\cdot\mid x_i,\hat{\vy}_{<t})\,\middle\|\,p_{\bar\theta}(\cdot\mid x_i,S_i,\hat{\vy}_{<t})\right).
    \label{eq:opsd}
\end{equation}
After each optimizer step, $\bar\theta \leftarrow \beta\bar\theta+(1-\beta)\theta$. The teacher therefore supplies a retrieval-conditioned view of the same model, while on-policy prefixes align supervision with the student's current state distribution. LoRA confines the update to a low-rank parameter subspace~\citep{hu2022lora}.

\section{Experiments}

\subsection{Experimental Setup}

\noindent \textbf{Data and domains.} We construct disjoint unlabeled training and test pools from three domains using SciKnowEval~\citep{feng2024sciknoweval} and MMLU-Pro~\citep{wang2024mmlupro}: mathematics (accounts for 20\%); natural science, including biology, chemistry, materials, and physics (each accounts for 10\%, 40\% in total); and humanities and social science, including psychology, history, economics, law, and global facts (each accounts for 8\%, 40\% in total). The test set consists of 500 new questions drawn from the same domains and in the same proportions. Gold labels are used only for evaluation.

\noindent \textbf{Models.} We study Qwen3-4B-Instruct~\citep{yang2025qwen3}, Qwen3.5-9B-Instruct~\citep{qwen3.5}, Ministral-3-8B-Instruct~\citep{liu2026ministral}, and Llama-3.1-8B-Instruct~\citep{llama2024herd}. Deploying the same protocol across these model families tests whether the selector depends on one representation geometry or capacity regime.

\noindent \textbf{Training protocol.} Unless varied in an ablation, every method uses 250 small-step optimizer updates (micro-batch size 4), a 20\% selection ratio, three retrieved demonstrations, five teacher candidates per target, and the same LoRA configuration and decoding budget. The EMA coefficient is $\beta=0.995$.

\noindent \textbf{Baselines.} \textsc{Original} is the unadapted model, and \textsc{Full} trains on all generated data without any additional selection. \textsc{Redundancy} implements the quality-diversity selection principle of DEITA~\citep{liu2024deita}. \textsc{Consistency} selects self-consistent teacher support motivated by SCOTT~\citep{wang2023scott} and PCSD~\citep{lv2026pcsd}. \textsc{NEURON} follows the internal-activation selection criterion of on-policy self-distillation~\citep{chen2026neuron}. \textsc{Mutual} adapts representation mutual-information maximization from chain-of-thought distillation~\citep{chen2024mutual}. SCOTT and Mutual originally use a stronger-teacher knowledge-distillation setting, and PCSD targets agentic reinforcement learning. We therefore re-implement their ideas inside our retrieval-conditioned self-distillation pipeline with identical data, generation, and update budgets. Detailed algorithm descriptions and fairness adjustments are provided in Appendix~\ref{app:baseline-details}.

\noindent \textbf{Metrics.} For every test question, we independently repeated the test three times. \emph{Avg@3} is the mean accuracy across the three rollout indices, and SD is their standard deviation. Table~\ref{tab:main} reports Avg@3 $\pm$ SD. \emph{Maj@3} is the accuracy of the majority-voted final answer. We also record end-to-end wall-clock time, including retrieval, teacher generation, selection, and optimization.

\subsection{Main Results}

\begin{table}[t]
\caption{Average rollout accuracy at 250 optimizer updates and a 20\% selection ratio. Each entry is Avg@3 $\pm$ SD (\%). The final row reports mean runtime per run. \textbf{Bold} and \underline{underlined} entries denote the best and second-best results, respectively. Maj@3 results are reported in Appendix~\ref{app:maj}.}
\label{tab:main}
\centering
\scriptsize
\setlength{\tabcolsep}{2.6pt}
\renewcommand{\arraystretch}{1.05}
\begin{tabularx}{\textwidth}{@{}ll*{7}{C}@{}}
\toprule
Model & Domain & Original & Full & Redundancy & Consistency & NEURON & Mutual & \method{} \\
\midrule
\multirow{4}{*}{\textit{Qwen3-4B}}
& Math & $45.3_{\pm 0.47}$ & $49.0_{\pm 0.82}$ & \underline{$49.7_{\pm 0.94}$} & $48.0_{\pm 1.41}$ & $46.7_{\pm 0.47}$ & $49.3_{\pm 1.70}$ & $\boldsymbol{51.3_{\pm 0.94}}$ \\
& Natural Sci. & $78.5_{\pm 0.82}$ & $78.8_{\pm 0.24}$ & $78.8_{\pm 0.62}$ & $78.7_{\pm 0.24}$ & $\boldsymbol{79.8_{\pm 0.85}}$ & \underline{$79.5_{\pm 1.08}$} & $79.2_{\pm 0.47}$ \\
& Hum. \& Soc. & $45.0_{\pm 0.71}$ & $43.8_{\pm 0.85}$ & \underline{$45.7_{\pm 1.03}$} & $45.2_{\pm 1.03}$ & $45.5_{\pm 0.71}$ & \underline{$45.7_{\pm 0.62}$} & $\boldsymbol{46.2_{\pm 0.47}}$ \\
\rowcolor{stdgray}& \textbf{Overall} & $58.5_{\pm 0.09}$ & $58.9_{\pm 0.34}$ & $59.7_{\pm 0.41}$ & $59.1_{\pm 0.52}$ & $59.5_{\pm 0.66}$ & \underline{$59.9_{\pm 0.90}$} & $\boldsymbol{60.4_{\pm 0.33}}$ \\
\midrule
\multirow{4}{*}{\textit{Qwen3.5-9B}}
& Math & $49.3_{\pm 0.47}$ & \underline{$51.3_{\pm 2.49}$} & \underline{$51.3_{\pm 1.70}$} & $49.0_{\pm 1.41}$ & $51.0_{\pm 1.41}$ & \underline{$51.3_{\pm 0.47}$} & $\boldsymbol{53.3_{\pm 2.05}}$ \\
& Natural Sci. & $80.2_{\pm 1.25}$ & \underline{$80.8_{\pm 1.25}$} & $80.7_{\pm 1.25}$ & $79.7_{\pm 0.47}$ & $\boldsymbol{81.7_{\pm 0.62}}$ & \underline{$80.8_{\pm 0.24}$} & $\boldsymbol{81.7_{\pm 0.62}}$ \\
& Hum. \& Soc. & $58.7_{\pm 0.94}$ & $\boldsymbol{61.2_{\pm 0.24}}$ & $57.2_{\pm 0.85}$ & $58.2_{\pm 1.03}$ & $57.7_{\pm 0.47}$ & \underline{$60.0_{\pm 0.71}$} & $58.5_{\pm 1.47}$ \\
\rowcolor{stdgray}& \textbf{Overall} & $65.4_{\pm 0.59}$ & $\boldsymbol{67.1_{\pm 1.06}}$ & $65.4_{\pm 0.57}$ & $64.9_{\pm 0.66}$ & $65.9_{\pm 0.19}$ & $66.6_{\pm 0.33}$ & \underline{$66.7_{\pm 0.77}$} \\
\midrule
\multirow{4}{*}{\textit{Ministral-3-8B}}
& Math & $42.3_{\pm 2.49}$ & $43.0_{\pm 0.82}$ & $44.0_{\pm 2.45}$ & $41.0_{\pm 0.82}$ & \underline{$45.3_{\pm 2.49}$} & $44.0_{\pm 1.41}$ & $\boldsymbol{46.7_{\pm 1.70}}$ \\
& Natural Sci. & $77.7_{\pm 1.65}$ & $78.2_{\pm 0.85}$ & $\boldsymbol{79.2_{\pm 0.24}}$ & $77.3_{\pm 1.03}$ & $78.3_{\pm 0.85}$ & $78.2_{\pm 1.93}$ & \underline{$78.5_{\pm 0.82}$} \\
& Hum. \& Soc. & $51.0_{\pm 1.08}$ & $\boldsymbol{57.0_{\pm 1.22}}$ & $55.3_{\pm 2.01}$ & $53.0_{\pm 2.55}$ & $54.7_{\pm 0.62}$ & $55.3_{\pm 0.85}$ & \underline{$56.0_{\pm 1.41}$} \\
\rowcolor{stdgray}& \textbf{Overall} & $59.9_{\pm 0.98}$ & \underline{$62.7_{\pm 0.50}$} & $62.6_{\pm 0.43}$ & $60.3_{\pm 1.09}$ & $62.3_{\pm 0.19}$ & $62.2_{\pm 1.07}$ & $\boldsymbol{63.1_{\pm 0.52}}$ \\
\midrule
\multirow{4}{*}{\textit{Llama-3.1-8B}}
& Math & $26.7_{\pm 3.30}$ & $25.0_{\pm 2.16}$ & $24.7_{\pm 1.25}$ & $27.7_{\pm 0.47}$ & $23.7_{\pm 2.62}$ & \underline{$28.7_{\pm 1.89}$} & $\boldsymbol{31.3_{\pm 1.25}}$ \\
& Natural Sci. & $78.2_{\pm 0.62}$ & \underline{$78.8_{\pm 0.62}$} & $78.2_{\pm 0.94}$ & $78.5_{\pm 0.41}$ & $\boldsymbol{79.0_{\pm 0.71}}$ & \underline{$78.8_{\pm 0.24}$} & $78.2_{\pm 0.24}$ \\
& Hum. \& Soc. & $39.2_{\pm 3.12}$ & $38.3_{\pm 1.93}$ & $37.7_{\pm 0.47}$ & $39.0_{\pm 0.82}$ & $38.8_{\pm 0.62}$ & $\boldsymbol{40.7_{\pm 1.31}}$ & \underline{$39.5_{\pm 1.08}$} \\
\rowcolor{stdgray} & \textbf{Overall} & $52.3_{\pm 2.04}$ & $51.9_{\pm 1.09}$ & $51.3_{\pm 0.52}$ & $52.5_{\pm 0.50}$ & $51.9_{\pm 0.98}$ & $\boldsymbol{53.5_{\pm 0.25}}$ & \underline{$53.3_{\pm 0.57}$} \\
\midrule
\multicolumn{2}{l}{Average Time Cost (min)} & 0.0 & 70.1 & 63.3 & 59.6 & 76.1 & 68.5 & 66.7 \\
\bottomrule
\end{tabularx}
\end{table}

\noindent\textbf{Effectiveness of our signal.}
Table~\ref{tab:main} first evaluates whether \method{} identifies more useful training data under a fixed update budget. Although it retains only 20\% of the unlabeled pool, \method{} outperforms \textsc{Full} on three of four backbones and raises their mean overall accuracy from 60.2\% to 60.9\%. It also achieves the highest cross-backbone average among the selection methods. The advantage is not uniform: \textsc{Full} remains better on Qwen3.5-9B and \textsc{Mutual} narrowly leads on Llama-3.1-8B. A post-hoc direction analysis provides complementary evidence for this selection signal: among the top 20\% of Qwen3-4B targets ranked by belief shift, \textbf{Fix} (the original answer is incorrect while the teacher's answer is correct) exceeds \textbf{Break} (correct to incorrect) by 13 percentage points and the retrieval-conditioned teacher accuracy rises from 23\% to 36\% (Appendix~\ref{app:belief-direction}).

\noindent\textbf{Generalization, efficiency, and domain dependence.}
The clearest pattern appears in mathematics. Across the four backbones, \method{} improves the corresponding unadapted checkpoints by 6.0, 4.0, 4.4, and 4.6 points. Gains are smaller in natural science, where the original checkpoints already achieve 77.7--80.2\%. This contrast suggests that the selection signal is most useful when retrieved evidence can revise an initially weak decision. \method{} requires 66.7 minutes per run, compared with 68.5 minutes for \textsc{Mutual} and 76.1 minutes for \textsc{NEURON}. Its performance therefore does not come from additional generation or optimization. In Section~\ref{sec:correction}, we further derive the convergence rate of the expected information gap during self-distillation with respect to optimization updates.

\noindent\textbf{Why self-distillation helps less on Llama.}
Llama-3.1-8B-Instruct exhibits weaker and less stable gains across the self-distillation methods in the long run (Figure~\ref{fig:updates}). In Table~\ref{tab:main}, \textsc{Full}, \textsc{Redundancy}, and \textsc{NEURON} all reduce its overall accuracy relative to the unadapted checkpoint, suggesting that the limitation is not specific to one selector. The direction analysis in Appendix~\ref{app:belief-direction} traces this pattern to the quality of the self-generated supervision. Within the top 20\% selected by belief shift, Qwen3-4B obtains 36\% Fix and 23\% Break, increasing teacher accuracy from 23\% to 36\%. The corresponding Llama subset obtains 26\% Fix and 27\% Break, while the teacher accuracy is slightly lower than the initial accuracy by 1.0 percentage point. Retrieval therefore produces a strong information change on Llama without satisfactory corrective outcomes. \method{} still improves its mathematics and overall accuracy by 4.6 and 1.0 points, respectively, but the near-zero Fix$-$Break balance helps explain why these gains remain unstable.

\subsection{Ablation Studies}

The following experiments separate the quality of the selection signal
from the effects of training duration and data volume.

\begin{figure}[t]
    \centering

    \begin{minipage}[t]{0.49\linewidth}
        \centering
        \includegraphics[width=\linewidth]{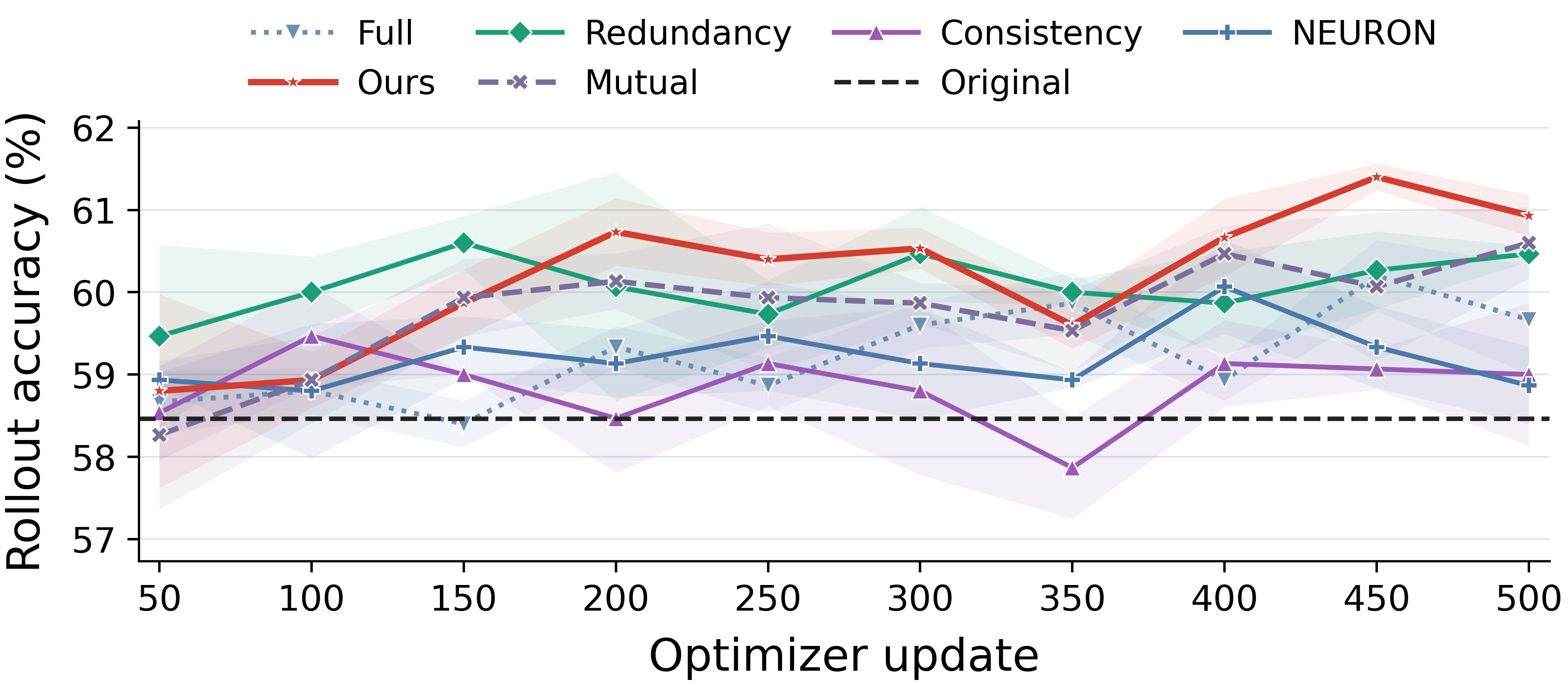}
        \par\smallskip
        \small(a) Qwen3-4B-Instruct
    \end{minipage}
    \hfill
    \begin{minipage}[t]{0.49\linewidth}
        \centering
        \includegraphics[width=\linewidth]{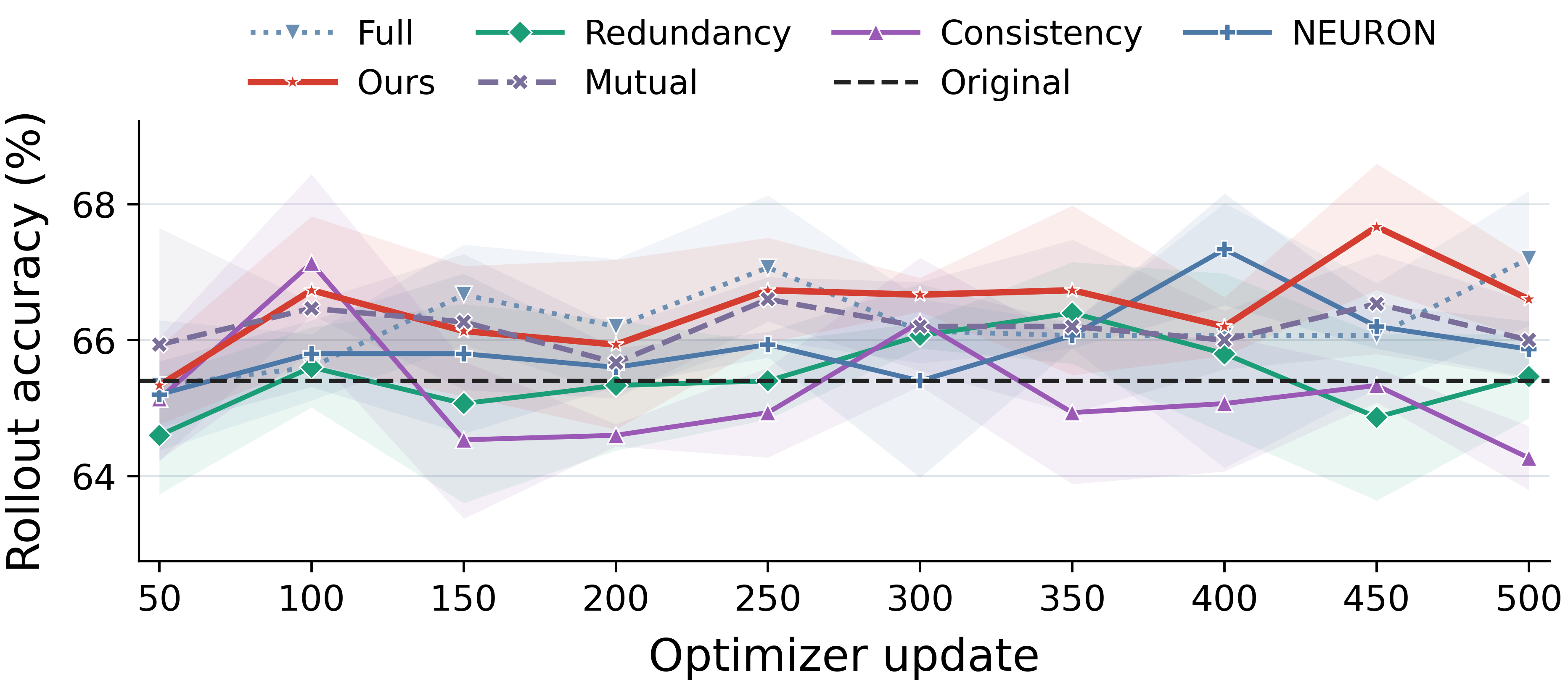}
        \par\smallskip
        \small(b) Qwen3.5-9B-Instruct
    \end{minipage}

    \vspace{0.5em}

    \begin{minipage}[t]{0.49\linewidth}
        \centering
        \includegraphics[width=\linewidth]{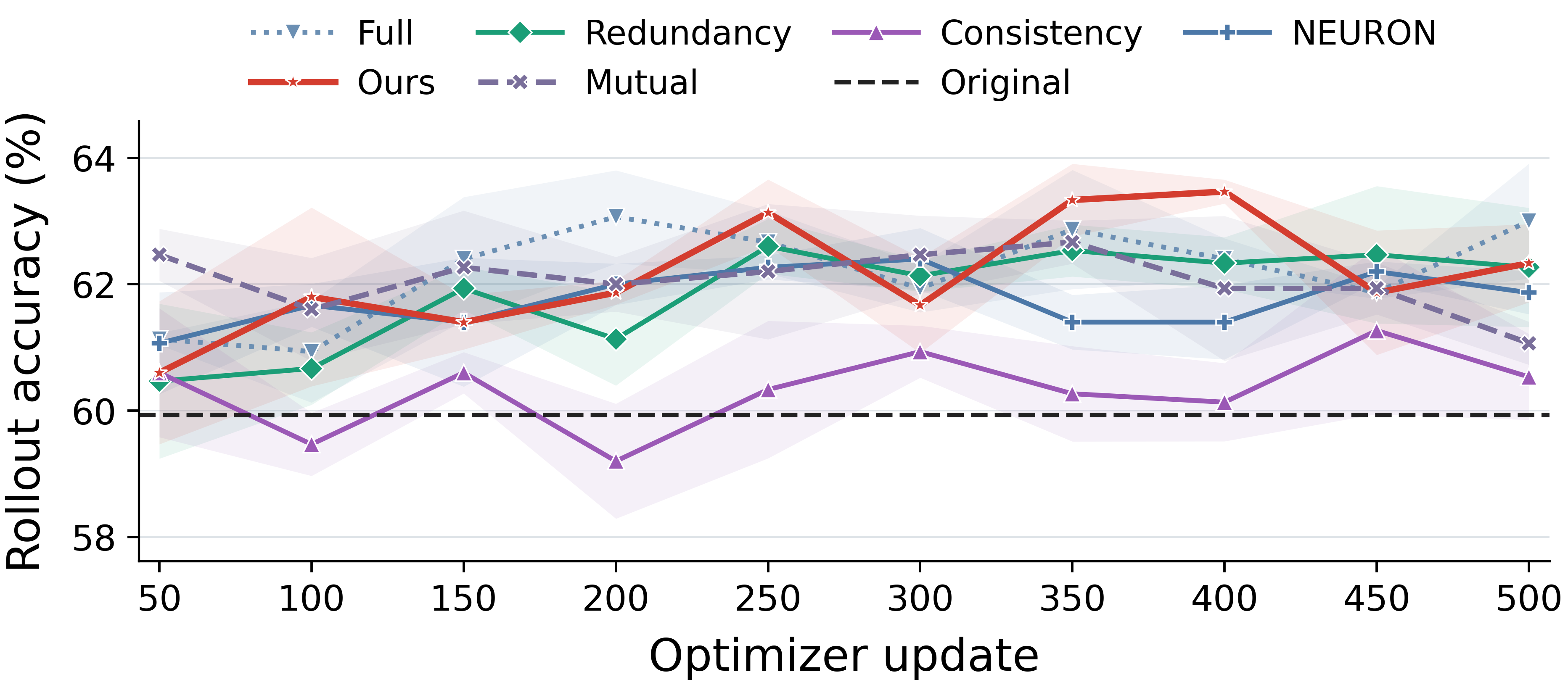}
        \par\smallskip
        \small(c) Ministral-3-8B-Instruct
    \end{minipage}
    \hfill
    \begin{minipage}[t]{0.49\linewidth}
        \centering
        \includegraphics[width=\linewidth]{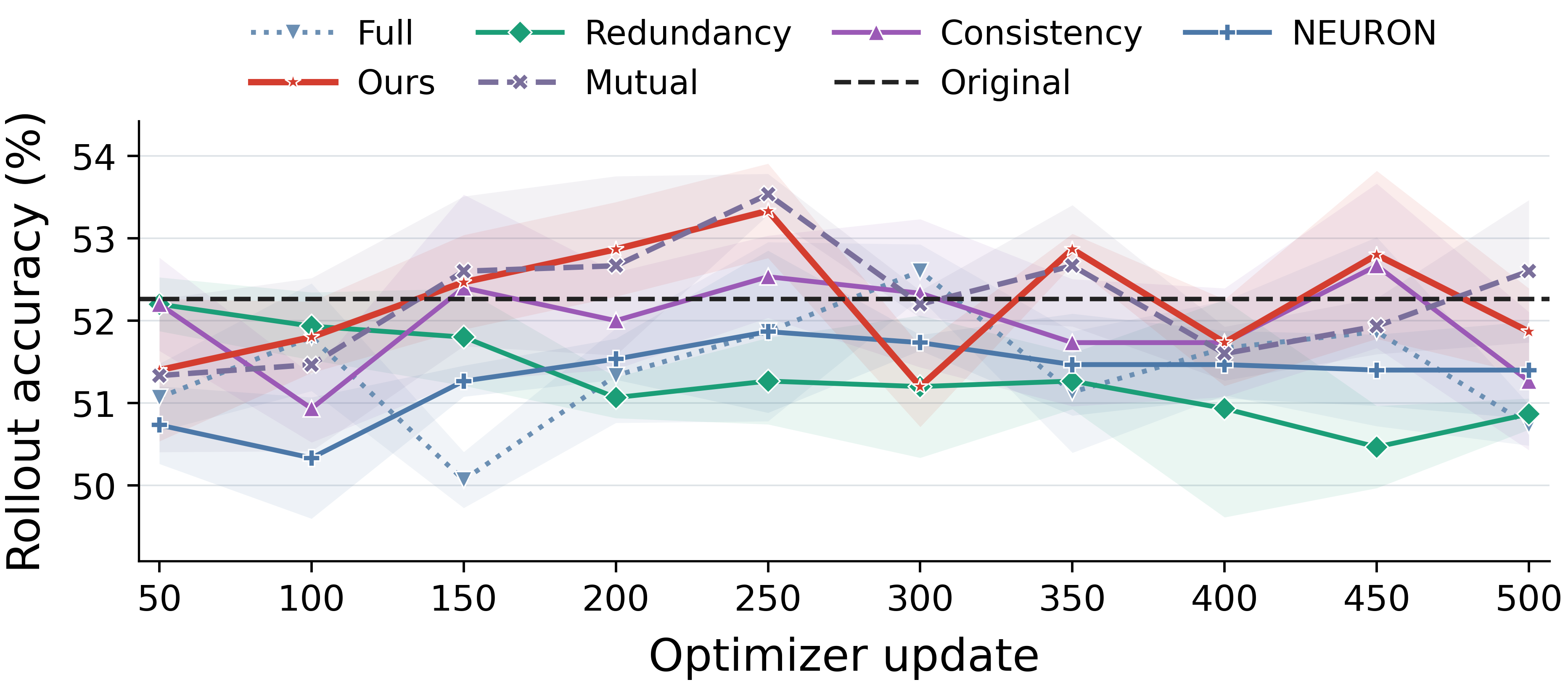}
        \par\smallskip
        \small(d) Llama-3.1-8B-Instruct
    \end{minipage}
    \caption{Accuracy over optimizer updates. Shaded bands denote the
    standard deviation across three rollout-index accuracies.}
    \label{fig:updates}
\end{figure}

\noindent\textbf{Sensitivity to the optimization budget.}
Figure~\ref{fig:updates} shows that self-distillation accuracy is
non-monotonic across update counts. \method{} remains above the
unadapted checkpoint over most of the trajectory on Qwen3-4B,
Qwen3.5-9B, and Ministral-3-8B, whereas the methods are less clearly
separated on Llama-3.1-8B. We use a common 250-update checkpoint for all
methods to avoid selecting model-specific peaks. These curves indicate
that data selection determines which supervision enters training, but
does not eliminate the backbone-dependent effects of prolonged fitting
to self-generated targets.

\begin{figure}[t]
    \centering

    \begin{minipage}[t]{0.485\linewidth}
        \centering
        \includegraphics[width=\linewidth]
        {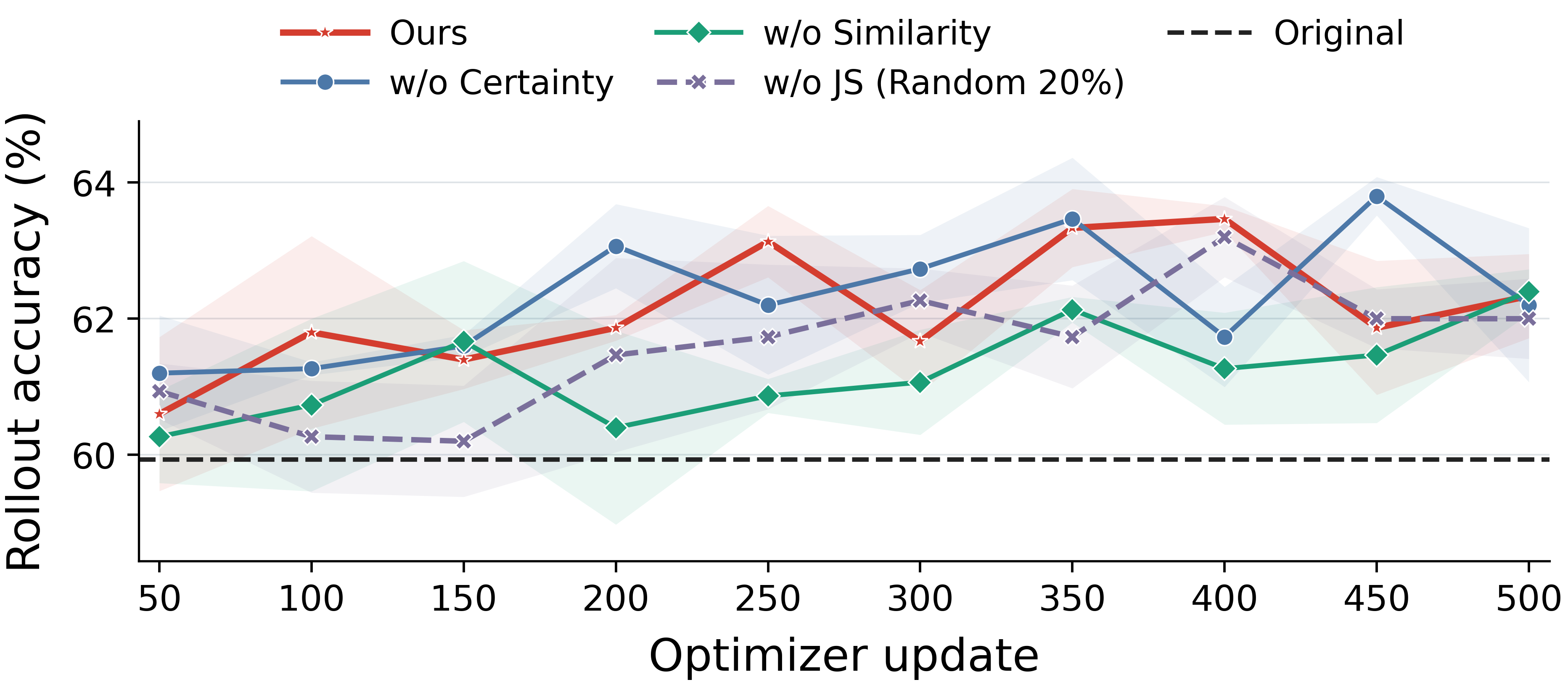}
        \vspace{-1em}
        \captionof{figure}{
            Component ablation on Ministral-3-8B-Instruct.
            Shaded bands denote SD across three rollout-index
            accuracies. ``w/o JS'' replaces belief-shift selection
            with a random 20\% subset; ``w/o Similarity'' and
            ``w/o Certainty'' remove the corresponding retrieval
            factors.
        }
        \label{fig:component-ablation}
    \end{minipage}
    \hfill
    \begin{minipage}[t]{0.485\linewidth}
        \centering
        \includegraphics[width=\linewidth]
        {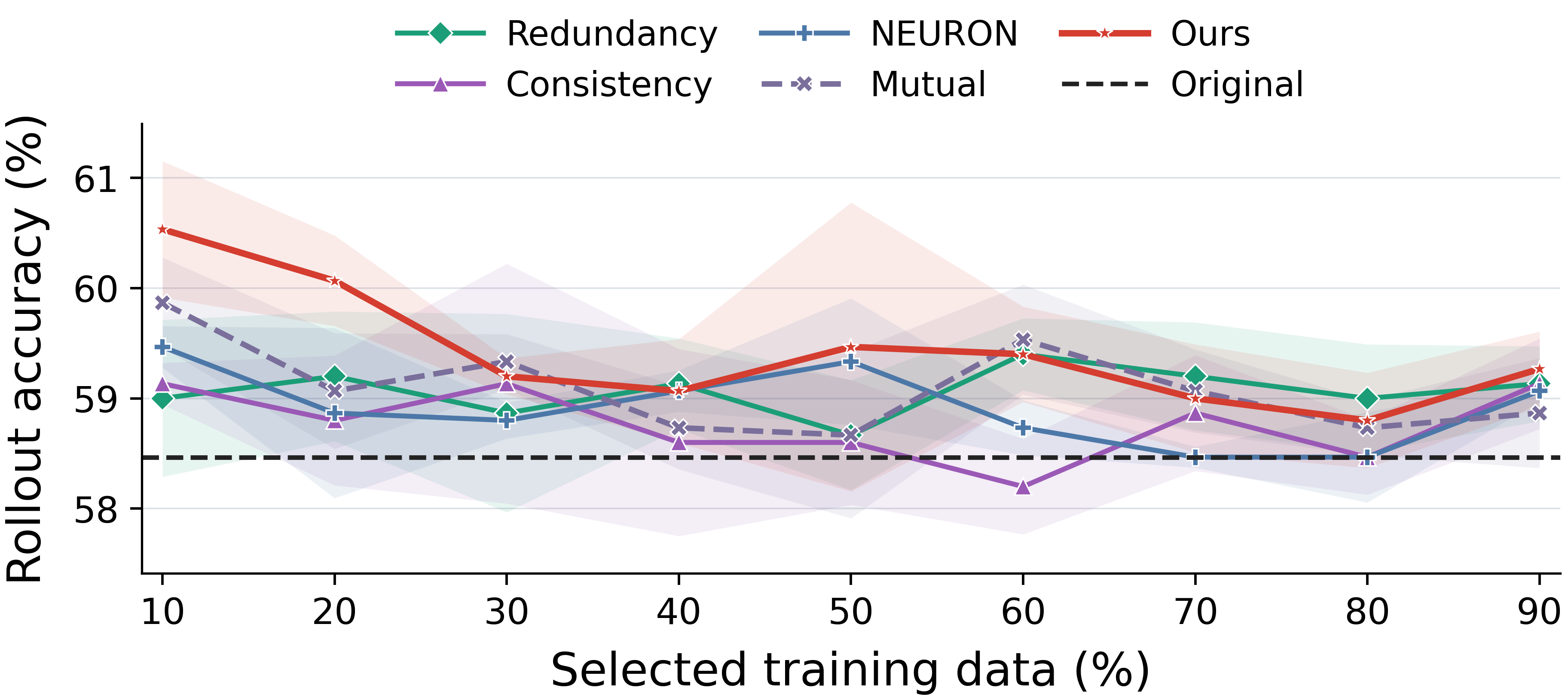}
        \vspace{-1em}
        \captionof{figure}{
            Selection-ratio ablation on Qwen3-4B-Instruct. Shaded bands denote SD across three rollout-index accuracies. The selected fraction of the unlabeled pool is varied while the generation and optimization budgets remain fixed.
        }
        \label{fig:selection-ratio}
    \end{minipage}

\end{figure}

\noindent\textbf{Does belief shift identify useful supervision?}
The most direct control is the ``w/o JS'' variant in
Figure~\ref{fig:component-ablation}: it preserves the same retrieval
procedure, selection ratio, and optimization budget, but replaces
belief-shift ranking with a random 20\% subset. At the default
250-update checkpoint, accuracy decreases from approximately 63.1\%
to 61.7\%. Since the retrieved contexts and amount of training data are
otherwise unchanged, this difference isolates the value of ranking
targets by their realized response to retrieval rather than merely
training on fewer examples.

Figure~\ref{fig:selection-ratio} tests the same claim across selection
budgets. \method{} has its clearest advantage at 10\% and 20\%, where
the selector must distinguish a small high-value subset from the
remaining pool. Its advantage contracts as the retained fraction
increases and lower-ranked targets enter training. This budget-dependent
pattern is evidence of a meaningful ordering: belief shift concentrates
useful self-distillation examples near the top of the ranking, rather
than benefiting all subsets equally. The balance between this
concentration and target coverage motivates the default ratio of 20\%.

\noindent\textbf{Which components make the signal informative?}
Figure~\ref{fig:component-ablation} further separates potential from
realized information. Removing trajectory similarity produces the
largest reduction at 250 updates, from approximately 63.1\% to 60.9\%,
showing that certainty alone cannot identify sources whose internal
decision process is relevant to the target. Removing certainty gives
62.2\%, while replacing JS selection with random selection gives
61.7\%. Similarity therefore supplies the main retrieval signal on
Ministral, certainty calibrates its reliability, and belief shift
determines which retrieval-conditioned targets enter distillation.
The occasional advantage of ``w/o Certainty'' at other checkpoints
also shows that certainty does not necessarily indicate correctness or effectiveness. For the motivation behind certainty calibration and further discussion, please refer to Section~\ref{sec: discussion}.

The retrieval-depth experiment is deferred to Appendix~\ref{app:retrieval-depth}, where Figure~\ref{fig:retrieval-k} tests the marginal benefit of increasing the amount of retrieved data without expanding the main-text ablation block.

\section{Discussion on the Theoretical Principles}
\label{sec: discussion}

\subsection{Maximizing Information Transfer for Self-Distillation}
\label{sec:information-analysis}

INFLOW organizes self-distillation as a two-stage information-transfer process. Retrieval determines how much information can be supplied to the teacher before generation, whereas data selection determines how much of the resulting information enters the distillation set. These decisions are made sequentially because the teacher's response to a context is unavailable when retrieval is performed.

For target $x_i$, let $\widetilde{\rho}_{ij}^{\,2}=\min\{c_js_{ij}^{2},1-\epsilon\}$ and $\gamma_{ij}=\widetilde{\rho}_{ij}^{\,2}/(1-\widetilde{\rho}_{ij}^{\,2})$. Under the local Gaussian channel model and conditionally independent source noises, the information supplied by a size-$k$ source set is
\begin{equation}
V_i(S)
=
I(T_i;Z_{iS})
=
\frac{1}{2}\log\left(1+\sum_{j\in S}\gamma_{ij}\right).
\label{eq:set-potential-information}
\end{equation}
Because Equation~\ref{eq:set-potential-information} is monotone in the sum of signal-to-noise ratios, retrieving the top-$k$ sources under $I_{ij}^{\mathrm{pot}}$ maximizes the information available to the target decision state before teacher generation.

After generation, let $C_i$ indicate whether the retrieved context $S_i$ is supplied, and let $A_i$ denote the resulting semantic answer. The realized information transferred to the teacher target is
\begin{equation}
I(A_i;C_i\mid X_i=x_i,S_i)
=
\operatorname{JS}(P_i^0\Vert P_i^{S_i})
=
B_i.
\label{eq:realized-transfer-information}
\end{equation}
Thus, $B_i$ measures in nats how much information about the retrieval intervention is expressed in the teacher's answer distribution.

\begin{proposition}[Information-transfer maximization]
\label{prop:information-transfer}
Assume that source observations are conditionally independent in the Gaussian channel and that teacher responses are conditionally independent across targets. Under retrieval depth $k$ and training-set size $m=\lceil\alpha N\rceil$, INFLOW satisfies
\begin{align}
S_i^\star
&\in
\arg\max_{|S|=k}
I(T_i;Z_{iS}),
\label{eq:optimal-potential-transfer}\\
\mathcal{T}^\star
&\in
\arg\max_{|\mathcal{T}|=m}
I(A_{\mathcal{T}};C_{\mathcal{T}}
\mid X_{\mathcal{T}},S_{\mathcal{T}}),
\label{eq:optimal-realized-transfer}
\end{align}
where $A_{\mathcal{T}}=(A_i)_{i\in\mathcal{T}}$ and
$C_{\mathcal{T}}=(C_i)_{i\in\mathcal{T}}$. Therefore, INFLOW maximizes the information supplied by retrieval and the total retrieval-induced information retained for distillation.
\end{proposition}

The two stages optimize different points of the same information path. Potential information is the largest amount supported by the internal retrieval channel before its effect can be observed. Realized information is the part that appears in the contextual teacher's prediction and is therefore available to the student through Equation~\ref{eq:opsd}. In contrast, \textsc{Mutual} uses representation-level conditional mutual information to measure information predicted before teacher generation, but does not verify how much of that information is expressed in the teacher targets. We actually considered  adapting mutual-information-guided knowledge distillation~\citep{chen2024mutual} to self-distillation as the core method, but ultimately found that the method derived in Proposition~\ref{prop:information-transfer} could make better use of the  information available in practice. The full proof is provided in Appendix~\ref{app:information-proofs}.
\subsection{Distillation Rate and Corrective Direction}
\label{sec:correction}

Let $q_i^S$ be the contextual teacher distribution and
$p_\theta$ the student sequence distribution on the selected targets.
The information gap optimized by on-policy distillation is
\begin{equation}
\mathcal{K}_{\mathcal{T}}(\theta)
=
\frac{1}{|\mathcal{T}|}
\sum_{i\in\mathcal{T}}
\operatorname{KL}
\left(
p_\theta(\cdot\mid x_i)
\,\Vert\,
q_i^S(\cdot\mid x_i)
\right).
\label{eq:distillation-information-gap}
\end{equation}
By the KL chain rule, this is the expected token-level objective in
Equation~\ref{eq:opsd}. Moreover, for the population semantic-answer
distributions induced by the same sequence models,
\begin{equation}
\operatorname{KL}(P_i^0\Vert P_i^S)
\le
\operatorname{KL}(p_{\theta_0}\Vert q_i^S),
\label{eq:data-processing-gap}
\end{equation}
so an answer-level change exposes part of the sequence-level information that distillation transfers.

\begin{proposition}[Convergence of the transferred information]
\label{prop:distillation-convergence}
Suppose the local distillation objective is $L$-smooth and satisfies the
Polyak--{\L}ojasiewicz condition with parameter $\mu$. For
$\eta\le 1/L$,
\begin{equation}
\mathcal{K}_{\mathcal{T}}(\theta_t)-\mathcal{K}_{\mathcal{T}}^\star
\le
(1-\mu\eta)^t
\left[
\mathcal{K}_{\mathcal{T}}(\theta_0)
-\mathcal{K}_{\mathcal{T}}^\star
\right].
\label{eq:distillation-convergence}
\end{equation}
\end{proposition}

A larger teacher--student gap supplies a stronger non-zero training signal, but requires more updates to reach the same absolute residual; the contraction rate itself is controlled by optimization geometry. More importantly, successful transfer does not imply correction. If $A_i^\star$ is the gold answer and $P_i^+$ is the post-distillation belief, then
\begin{equation}
\frac{1}{|\mathcal{T}|}\sum_{i\in\mathcal{T}}
\left[P_i^+(A_i^\star)-P_i^0(A_i^\star)\right]
\ge
\Delta_{\mathcal{T}}-E_{\mathcal{T}},
\label{eq:correction-bound}
\end{equation}
where
$\Delta_{\mathcal{T}}
=
\frac{1}{|\mathcal{T}|}\sum_{i\in\mathcal{T}}
\left[P_i^S(A_i^\star)-P_i^0(A_i^\star)\right],
\qquad
E_{\mathcal{T}}
=
\frac{1}{|\mathcal{T}|}\sum_{i\in\mathcal{T}}
\operatorname{TV}(P_i^+,P_i^S).$

The convergence of the teacher--student information gap measures how faithfully contextual information is transferred, but does not guarantee correct-label repair: as Appendix~\ref{app:belief-direction} shows, a large information shift may instead preserve or introduce an erroneous target. For Llama-3.1-8B, its low initial accuracy implies that the unlabeled source pool contains many incorrect model answers, yielding a small or negative $\Delta_{\mathcal{T}}$ and propagating errors through distillation. \method{} improves overall accuracy by 1.0 percentage point after 250 updates yet declines after 500 updates, suggesting that further erasure of information gaps in distillation is not necessarily beneficial.
\section{Conclusion}

We propose \method{} to connect two decisions that previous annotation-free self-distillation methods largely treat separately: which evidence should condition the teacher and which resulting targets should be retained for training. \method{} formulates this process as budgeted information-flow optimization. Certainty-calibrated retrieval maximizes the information available to the target under a local channel model, while semantic belief shift allocates the training budget according to the information expected to be transferred. Our analysis further relates the induced answer-level gap to the sequence-level distillation objective while acknowledging that information gaps in annotation-free self-distillation cannot fully guide the correction of incorrect answers. Experiments across four model backbones, together with posterior statistics, ablation studies, and hyperparameter analysis, support \method{} as an effective data-selection signal. Overall, \method{} provides a unified basis for guiding knowledge retrieval and data selection in self-distillation through information flow optimization.

\section*{AI Use Statement}

Generative AI tools, primarily OpenAI Codex, were used to assist with research framing, feedback, manuscript drafting, and language editing. All AI-assisted text was reviewed by the authors. The authors take full responsibility for the accuracy, originality, and final content of this work.

\section*{Ethics Statement}

This work uses publicly available academic benchmarks, and does not involve newly collected personal data or human-subject experiments. The datasets and models are used in accordance with their stated terms and licenses. Annotation-free self-distillation can reproduce or amplify incorrect and biased predictions already present in a model, particularly when confident but erroneous outputs are selected as supervision. INFLOW measures whether retrieved information changes the teacher's belief, but does not by itself certify that the change is factually correct or socially unbiased. The method should therefore be combined with appropriate validation and safety evaluation before deployment in consequential applications.

\section*{Reproducibility Statement}

The method section specifies the trajectory representation, certainty-calibrated retrieval score, semantic belief construction, belief-shift criterion, and on-policy distillation objective; Algorithm~\ref{alg:inflow} summarizes the complete training procedure. The experimental setup reports the evaluated models and datasets, generation settings, retrieval and selection budgets, optimization configuration, and evaluation protocol. The appendix provides the prompts, baseline implementations, additional hyperparameter analyses, qualitative retrieval examples, and case studies. Together, these materials are intended to support independent reproduction of the method and reported experiments.

\bibliographystyle{plainnat}
\bibliography{references}

\clearpage
\appendix
\appendix

\section{Algorithmic Description}
\subsection{Algorithm pseudocode}
\label{app:algorithm}

Algorithm~\ref{alg:inflow} summarizes the complete procedure, from internal-state extraction and retrieval to belief-shift selection and on-policy optimization.

\begin{algorithm}[t]
\caption{\method{} On-Policy Self-Distillation}
\label{alg:inflow}
\begin{algorithmic}[1]
\Require Unlabeled questions $\mathcal{U}=\{x_i\}_{i=1}^{N}$; model $p_\theta$; retrieval depth $k$; teacher samples $M$; selection ratio $\alpha$; belief temperature $T$; EMA coefficient $\beta$; optimizer $\operatorname{Opt}$; optimizer updates $J$; minibatch size $b$
\Ensure Adapted model $p_\theta$
\State Initialize EMA teacher parameters $\bar{\theta}\gets\theta$

\Statex \textbf{Stage I: Estimate potential information}
\For{$x_i\in\mathcal{U}$}
    \State Generate an initial response with $p_{\bar{\theta}}$
    \State Extract the layerwise final-answer trajectory $\tau_i$
    \State Compute the legal-answer belief $q_i$ at the fixed answer-scoring position
    \State $c_i\gets 1-H(q_i)/\log|\mathcal{A}_i|$
\EndFor
\For{$x_i\in\mathcal{U}$}
    \For{$x_j\in\mathcal{U}\setminus\{x_i\}$}
        \State Compute $I_{ij}^{\mathrm{pot}}$ from $\tau_i$, $\tau_j$, and $c_j$
    \EndFor
    \State $S_i\gets\operatorname{TopK}_{j\neq i}\!\left(I_{ij}^{\mathrm{pot}},k\right)$
\EndFor

\Statex \textbf{Stage II: Measure realized information}
\For{$x_i\in\mathcal{U}$}
    \State Compute $q_i^{0}$ using Equation~\ref{eq:legal-answer-belief}
    \State Sample $\{r_i^{S,m}\}_{m=1}^{M}\sim p_{\bar{\theta}}(\cdot\mid x_i,S_i)$
    \For{$m=1,\ldots,M$}
        \State Compute $q_i^{S,m}$ using Equation~\ref{eq:legal-answer-belief}
    \EndFor
    \State Estimate $P_i^0$ and $P_i^S$ using Equation~\ref{eq:semantic-belief}
    \State $B_i\gets\operatorname{JS}\!\left(P_i^0\Vert P_i^{S}\right)$
\EndFor
\State $\mathcal{D}\gets\operatorname{TopFraction}_{i}\!\left(\{(x_i,S_i)\}_{i=1}^{N},B_i,\alpha\right)$
\label{alg:select}

\Statex \textbf{Stage III: On-policy self-distillation}
\For{$j=1,\ldots,J$}
    \State Sample a minibatch $\mathcal{B}_j\subset\mathcal{D}$ with $|\mathcal{B}_j|=b$
    \State $\mathcal{L}_j\gets 0$
    \For{$(x_i,S_i)\in\mathcal{B}_j$}
        \State Sample $\hat{y}_i\sim p_\theta(\cdot\mid x_i)$
        \State $\ell_i\gets 0$
        \For{$t=1,\ldots,|\hat{y}_i|$}
            \State $\displaystyle
            \ell_i\gets\ell_i+
            \operatorname{KL}\!\left(
            p_\theta(\cdot\mid x_i,\hat{y}_{i,<t})
            \,\Vert\,
            p_{\bar{\theta}}(\cdot\mid x_i,S_i,\hat{y}_{i,<t})
            \right)$
        \EndFor
        \State $\mathcal{L}_j\gets\mathcal{L}_j+\ell_i/|\hat{y}_i|$
    \EndFor
    \State $\mathcal{L}_j\gets\mathcal{L}_j/|\mathcal{B}_j|$
    \State $\theta\gets\operatorname{Opt}\!\left(\theta,\nabla_\theta\mathcal{L}_j\right)$
    \State $\bar{\theta}\gets\beta\bar{\theta}+(1-\beta)\theta$
\EndFor
\State \Return $p_\theta$
\end{algorithmic}
\end{algorithm}

\subsection{Semantic Beliefs for Open-Ended Generation}
\label{app:free-form-belief}

For tasks without a predefined answer set, answer-token probabilities cannot be normalized over a shared collection of legal options. We instead construct a discrete semantic space from sampled responses, following the semantic-entropy principle~\citep{kuhn2023semantic,farquhar2024detecting}. Let $r_i^{0,1}$ denote the initial response and let $\{r_i^{S,m}\}_{m=1}^{M}$ denote the retrieval-conditioned teacher responses. We cluster the union
\begin{equation}
\mathcal{R}_i
=
\{r_i^{0,1}\}
\cup
\{r_i^{S,m}\}_{m=1}^{M}
\end{equation}
into semantic classes $\mathcal{G}_i=\{G_{i,1},\ldots,G_{i,K_i}\}$. Two responses are assigned to the same class when a natural-language-inference model predicts entailment in both directions. This criterion merges surface-distinct generations that express the same answer while separating responses with incompatible conclusions.

Let $g_i(r)\in\mathcal{G}_i$ denote the semantic class assigned to response $r$. For each retrieval-conditioned generation, we compute its length-normalized sequence score
\begin{equation}
\ell_i^{S,m}
=
\frac{1}{|r_i^{S,m}|}
\sum_{t=1}^{|r_i^{S,m}|}
\log p_{\bar{\theta}}
\left(
r_{i,t}^{S,m}
\mid
x_i,S_i,r_{i,<t}^{S,m}
\right)
\end{equation}
and normalize the scores across the sampled responses:
\begin{equation}
w_i^{S,m}
=
\frac{
\exp\!\left(\ell_i^{S,m}/T_s\right)
}{
\sum_{m'=1}^{M}
\exp\!\left(\ell_i^{S,m'}/T_s\right)
}.
\label{eq:free-form-response-weight}
\end{equation}
The initial and retrieval-conditioned beliefs over semantic classes are then
\begin{equation}
P_i^0(G)
=
\mathbb{I}\!\left[g_i(r_i^{0,1})=G\right],
\qquad
P_i^S(G)
=
\sum_{m=1}^{M}
w_i^{S,m}
\mathbb{I}\!\left[g_i(r_i^{S,m})=G\right].
\label{eq:free-form-semantic-belief}
\end{equation}
When sequence scores are unavailable, setting $w_i^{S,m}=1/M$ reduces Equation~\ref{eq:free-form-semantic-belief} to the empirical frequency of each semantic class. Because both distributions are defined on the shared class set $\mathcal{G}_i$, their retrieval-induced shift is computed using the same Jensen--Shannon divergence as in Equation~\ref{eq:belief-shift}.

\section{Proofs for the Information Analysis}
\label{app:information-proofs}

\subsection{Proof of Proposition~\ref{prop:information-transfer}}
\label{app:proof-information-transfer}

Fix target $i$. Under the local Gaussian model, an invertible rescaling of the standardized channel in Equation~\ref{eq:local-channel} gives
\begin{equation}
R_{ij}
=
\sqrt{\gamma_{ij}}T_i+\varepsilon_{ij},
\qquad
T_i\sim\mathcal{N}(0,1),
\qquad
\varepsilon_{ij}\overset{\mathrm{i.i.d.}}{\sim}\mathcal{N}(0,1).
\label{eq:gaussian-source-channel}
\end{equation}
For a source set $S$, write
\[
R_{iS}=v_ST_i+\varepsilon_S,
\qquad
v_S=(\sqrt{\gamma_{ij}})_{j\in S}.
\]
The conditional covariance of $R_{iS}$ given $T_i$ is $I$, while its marginal covariance is $I+v_Sv_S^\top$. Gaussian entropy and the matrix determinant lemma yield
\begin{align}
I(T_i;R_{iS})
&=
\frac{1}{2}\log\det(I+v_Sv_S^\top)\\
&=
\frac{1}{2}\log(1+v_S^\top v_S)\\
&=
\frac{1}{2}\log\left(1+\sum_{j\in S}\gamma_{ij}\right).
\end{align}
Since the logarithm is increasing, the maximizing size-$k$ set contains the $k$ largest $\gamma_{ij}$. Moreover,
\[
I_{ij}^{\mathrm{pot}}
=
\frac{1}{2}\log(1+\gamma_{ij})
=
-\frac{1}{2}\log(1-\widetilde{\rho}_{ij}^{\,2}),
\]
so ranking by $I_{ij}^{\mathrm{pot}}$ produces the same set. This proves Equation~\ref{eq:optimal-potential-transfer}.

For the realized stage, let
$C_i\sim\operatorname{Bernoulli}(1/2)$, with
\[
p(A_i\mid C_i=0)=P_i^0,
\qquad
p(A_i\mid C_i=1)=P_i^{S_i}.
\]
The marginal answer distribution is
$M_i=(P_i^0+P_i^{S_i})/2$. Hence
\begin{align}
I(A_i;C_i\mid X_i=x_i,S_i)
&=
\frac{1}{2}\operatorname{KL}(P_i^0\Vert M_i)
+
\frac{1}{2}\operatorname{KL}(P_i^{S_i}\Vert M_i)\\
&=
\operatorname{JS}(P_i^0\Vert P_i^{S_i})\\
&=
B_i.
\end{align}

Conditional independence across targets gives
\begin{align}
I(A_{\mathcal{T}};C_{\mathcal{T}}
\mid X_{\mathcal{T}},S_{\mathcal{T}})
&=
\sum_{i\in\mathcal{T}}
I(A_i;C_i\mid X_i=x_i,S_i)\\
&=
\sum_{i\in\mathcal{T}}B_i.
\label{eq:joint-transfer-decomposition}
\end{align}
For fixed $|\mathcal{T}|=m$, maximizing the joint information in Equation~\ref{eq:joint-transfer-decomposition} is therefore equivalent to selecting the $m$ largest $B_i$. This is precisely the semantic belief-shift selection rule and proves Equation~\ref{eq:optimal-realized-transfer}.

\subsection{Why Representation CMI Differs from Realized Flow}
\label{app:mutual-flow-difference}

For \textsc{Mutual}, let $S_i^{(0)}=\varnothing$ and greedily select
\begin{equation}
j_t^\star
=
\arg\max_{j\notin S_i^{(t-1)}\cup\{i\}}
\widehat I
\left(
\vz_i;\vz_j
\mid
\vz_{S_i^{(t-1)}}
\right),
\qquad
S_i^{(t)}=S_i^{(t-1)}\cup\{j_t^\star\}.
\label{eq:mutual-greedy-proof}
\end{equation}
Its accumulated retrieval score is
\begin{equation}
U_i^{\mathrm{mut}}
=
\sum_{t=1}^{k}
\widehat I
\left(
\vz_i;\vz_{j_t^\star}
\mid
\vz_{S_i^{(t-1)}}
\right)
\approx
\widehat I(\vz_i;\vz_{S_i^{(k)}}).
\label{eq:mutual-accumulated-proof}
\end{equation}
This quantity estimates non-redundant dependence among the target and
source representations. It contains neither the contextual answer
distribution $P_i^S$ nor the realized flow $b_i(S)$. Consequently,
$U_i^{\mathrm{mut}}$ remains unchanged whether the contextual teacher
uses or ignores the retrieved sources.

Both methods require a pre-generation proxy to avoid generating under
all possible contexts. Their difference appears in target selection.
\textsc{Mutual} ranks targets using $U_i^{\mathrm{mut}}$, so an
end-to-end guarantee would require its representation score to preserve
realized-flow ordering among candidate contexts and across
different targets. INFLOW uses the pre-generation proxy only within
each target's retrieval problem and then ranks targets by the measured
$\widehat B_i(\widehat S_i)$. Proposition~\ref{prop:information-transfer}
therefore separates its approximation error into within-target
retrieval calibration and finite-sample belief estimation.

Neither quantity determines whether the intervention is correct.
Representation CMI measures internal statistical dependence, while
$B_i(S)$ measures the teacher's answer-level response to retrieval.
Connecting either quantity to
$I(A_i^\star;Z_{iS}\mid X_i)$ would require an additional assumption
relating model beliefs to the unknown gold answer.

\subsection{Proof of Proposition~\ref{prop:distillation-convergence}}

For each selected target, let
$p_\theta(y\mid x_i)$ and $q_i^S(y\mid x_i)$ be the student and
retrieval-conditioned teacher sequence distributions. Autoregressive
factorization and the chain rule of KL divergence give
\begin{align}
&\operatorname{KL}
\left(
p_\theta(\cdot\mid x_i)
\Vert
q_i^S(\cdot\mid x_i)
\right)\\
&\quad=
\mathbb{E}_{y\sim p_\theta(\cdot\mid x_i)}
\sum_{r=1}^{|y|}
\operatorname{KL}
\left(
p_\theta(\cdot\mid x_i,y_{<r})
\Vert
q_i^S(\cdot\mid x_i,y_{<r})
\right).
\label{eq:sequence-kl-chain-rule}
\end{align}
Averaging Equation~\ref{eq:sequence-kl-chain-rule} over
$\mathcal{T}$ yields Equation~\ref{eq:distillation-information-gap}.

Let $g(y)$ map a complete sequence to its semantic answer. Since
deterministic post-processing cannot increase KL divergence,
\begin{equation}
\operatorname{KL}
\left(
g_\#p_{\theta_0}
\Vert
g_\#q_i^S
\right)
\le
\operatorname{KL}
\left(
p_{\theta_0}
\Vert
q_i^S
\right),
\end{equation}
where $g_\#$ denotes the induced answer distribution. Substituting
$g_\#p_{\theta_0}=P_i^0$ and $g_\#q_i^S=P_i^S$ proves
Equation~\ref{eq:data-processing-gap}.

Now abbreviate $\mathcal{K}(\theta)$ by $\mathcal{K}_\theta$.
$L$-smoothness and the update
$\theta_{t+1}=\theta_t-\eta\nabla\mathcal{K}_{\theta_t}$ imply
\begin{align}
\mathcal{K}_{\theta_{t+1}}
&\le
\mathcal{K}_{\theta_t}
-
\eta\left(1-\frac{L\eta}{2}\right)
\|\nabla\mathcal{K}_{\theta_t}\|_2^2\\
&\le
\mathcal{K}_{\theta_t}
-
\frac{\eta}{2}
\|\nabla\mathcal{K}_{\theta_t}\|_2^2,
\end{align}
where the second inequality uses $\eta\le1/L$. Under the
Polyak--{\L}ojasiewicz condition,
\begin{equation}
\frac{1}{2}
\|\nabla\mathcal{K}_{\theta}\|_2^2
\ge
\mu\bigl(
\mathcal{K}(\theta)-\mathcal{K}^\star
\bigr).
\end{equation}
For exact local teacher fitting gives
\begin{equation}
\mathcal{K}(\theta_{t+1})
\le
(1-\mu\eta)\bigl(
\mathcal{K}(\theta)-\mathcal{K}^\star
\bigr).
\end{equation}
Iterating proves Equation~\ref{eq:distillation-convergence}. Solving
$(1-\mu\eta)^t\mathcal{K}(\theta_0)\le\varepsilon$ for $t$ proves
Equation~\ref{eq:distillation-convergence}.

\subsection{Effect of EMA Teacher Drift}

Proposition~\ref{prop:distillation-convergence} treats the teacher as locally
fixed. Let $\mathcal{K}_t$ denote the objective induced by the EMA
teacher at update $t$, and suppose its one-step drift satisfies
\begin{equation}
\mathcal{K}_{t+1}(\theta)
\le
\mathcal{K}_t(\theta)+\delta_t
\qquad
\text{for all local }\theta.
\end{equation}
Applying the previous contraction before each teacher update yields
\begin{equation}
\mathcal{K}_t(\theta_t)
\le
(1-\mu\eta)^t\mathcal{K}_0(\theta_0)
+
\sum_{r=0}^{t-1}
(1-\mu\eta)^{t-1-r}\delta_r.
\label{eq:ema-tracking}
\end{equation}
Thus EMA drift produces a tracking floor rather than changing the
local geometric rate. A slowly moving teacher has small $\delta_t$;
large or inconsistent contextual changes can prevent the student from
closing the information gap within the fixed update budget.

\subsection{Direction of the Teacher Information}

Let
\[
\hat A_i^0=\arg\max_aP_i^0(a),
\qquad
\hat A_i^S=\arg\max_aP_i^S(a).
\]
The change in teacher hard-answer accuracy is
\begin{align}
&\frac{1}{|\mathcal{T}|}
\sum_{i\in\mathcal{T}}
\left(
\mathbb{I}[\hat A_i^S=A_i^\star]
-
\mathbb{I}[\hat A_i^0=A_i^\star]
\right)\\
&\qquad=
\frac{N_{\mathrm{fix}}-N_{\mathrm{break}}}
{|\mathcal{T}|}.
\end{align}
A Fix contributes $+1$, a Break contributes $-1$, and unchanged or
wrong-to-wrong transitions contribute zero.

This direction is not determined by the information gap. For any
$P_i^0\neq P_i^S$, at least one answer gains probability and another
loses probability. Assigning either answer as $A_i^\star$ preserves
$B_i$ and the teacher--student KL while reversing whether the same
distillation target is beneficial. Consequently, rapid convergence of
Equation~\ref{eq:distillation-convergence} guarantees faithful transfer
of the contextual teacher distribution, not correction of its answer.

\section{Additional Experimental Results}
\subsection{Majority-Vote Results}
\label{app:maj}

\begin{table}[t]
\caption{3-rollout Majority-vote accuracy (Maj@3, \%) at 250 optimizer updates and a 20\% selection ratio. \textbf{Bold} and \underline{underlined} values are the best and second best in each row. In the event of a tie, the voting outcome is determined by the softmax of the average answer token logit.}
\label{tab:maj-results}
\centering
\scriptsize
\setlength{\tabcolsep}{2.8pt}
\renewcommand{\arraystretch}{1.05}
\begin{tabularx}{\textwidth}{@{}ll*{7}{C}@{}}
\toprule
Model & Domain & Original & Full & Redundancy & Consistency & NEURON & Mutual & \method{} \\
\midrule
\multirow{4}{*}{\textit{Qwen3-4B}}
& Math & 47.0 & 50.0 & \underline{51.0} & 50.0 & 46.0 & 50.0 & \textbf{52.0} \\
& Natural Sci. & 78.5 & \underline{79.5} & 78.5 & 78.5 & \textbf{80.0} & \underline{79.5} & 79.0 \\
& Hum. \& Soc. & 44.5 & 44.0 & 45.0 & \textbf{46.0} & 45.0 & \underline{45.5} & \textbf{46.0} \\
\rowcolor{stdgray}& \textbf{Overall} & 58.6 & 59.4 & 59.6 & 59.8 & 59.2 & \underline{60.0} & \textbf{60.4} \\
\midrule
\multirow{4}{*}{\textit{Qwen3.5-9B}}
& Math & \underline{53.0} & \underline{53.0} & \textbf{54.0} & 49.0 & \textbf{54.0} & 50.0 & \textbf{54.0} \\
& Natural Sci. & 80.5 & 80.0 & 81.0 & \underline{82.0} & \textbf{82.5} & 81.0 & \textbf{82.5} \\
& Hum. \& Soc. & \textbf{61.0} & \underline{60.5} & 56.5 & 59.0 & 58.0 & 59.0 & 60.0 \\
\rowcolor{stdgray}& \textbf{Overall} & \underline{67.2} & 66.8 & 65.8 & 66.2 & 67.0 & 66.0 & \textbf{67.8} \\
\midrule
\multirow{4}{*}{\textit{Ministral-3-8B}}
& Math & 42.0 & 43.0 & 43.0 & 40.0 & \underline{44.0} & 43.0 & \textbf{47.0} \\
& Natural Sci. & 78.0 & \underline{80.0} & \textbf{80.5} & 77.0 & 78.5 & 79.0 & 79.5 \\
& Hum. \& Soc. & 51.5 & \textbf{57.5} & 56.0 & 54.0 & \underline{56.5} & \textbf{57.5} & 55.5 \\
\rowcolor{stdgray}& \textbf{Overall} & 60.2 & \textbf{63.6} & 63.2 & 60.4 & 62.8 & 63.2 & \underline{63.4} \\
\midrule
\multirow{4}{*}{\textit{Llama-3.1-8B}}
& Math & \underline{30.0} & 26.0 & 22.0 & 28.0 & 24.0 & 28.0 & \textbf{31.0} \\
& Natural Sci. & \textbf{79.5} & \underline{79.0} & 78.5 & 78.0 & \underline{79.0} & \underline{79.0} & 78.5 \\
& Hum. \& Soc. & 38.0 & 38.0 & 38.0 & \underline{40.0} & 39.5 & \textbf{41.0} & \underline{40.0} \\
\rowcolor{stdgray}& \textbf{Overall} & \underline{53.0} & 52.0 & 51.0 & 52.8 & 52.2 & \textbf{53.6} & \textbf{53.6} \\
\bottomrule
\end{tabularx}
\end{table}

Majority voting preserves the cross-model conclusion from Table~\ref{tab:main}. \method{} reaches the best or tied-best overall Maj@3 on Qwen3-4B, Qwen3.5-9B, and Llama-3.1-8B, and is 0.2 points behind Full on Ministral. Averaged across the four backbones, its overall majority-vote accuracy is 61.3\%, compared with 60.7\% for Mutual, the strongest baseline average, and 59.8\% for the unadapted models. The gains over Original are 1.8, 0.6, 3.2, and 0.6 points. The smaller Qwen3.5 gain after voting is consistent with its high unadapted Maj@3, while the Ministral result shows that broader Full supervision can improve answer consensus even when \method{} has the higher average-rollout score.

\subsection{Direction of Retrieval-Induced Belief Shifts}
\label{app:belief-direction}

We examine whether retrieval-induced belief shifts tend to produce corrective supervision on Qwen3-4B and Llama-3.1-8B. For each model, all targets are ranked by $B_i$ in descending order and divided into ten equally sized intervals. The initial and retrieval-conditioned decisions are defined as $\arg\max_a P_i^0(a)$ and $\arg\max_a P_i^S(a)$, respectively. A \emph{Fix} changes an incorrect initial prediction to the gold answer, a \emph{Break} changes a correct prediction to an incorrect one, and the remaining cases are categorized as wrong-to-wrong or right-to-right. Gold labels are used only in this post-hoc analysis.

\begin{table}[t]
\caption{Direction of retrieval-induced belief shifts on Qwen3-4B. Targets are ranked by JS belief shift from largest to smallest. Transition frequencies and accuracies are percentages within each interval; Fix$-$Break is measured in percentage points.}
\label{tab:belief-direction-qwen}
\centering
\setlength{\tabcolsep}{3.2pt}
\renewcommand{\arraystretch}{1.06}
\begin{tabular}{lrrrrrrrr}
\toprule
JS rank
& Fix
& Break
& W$\rightarrow$W
& R$\rightarrow$R
& Fix$-$Break
& Initial Acc.
& Teacher Acc.
& Mean JS \\
\midrule
0--10\%   & 36.0 & 30.0 & 34.0 &  0.0 &  6.0 & 30.0 & 36.0 & 0.6931 \\
10--20\%  & 36.0 & 16.0 & 48.0 &  0.0 & 20.0 & 16.0 & 36.0 & 0.5898 \\
20--30\%  & 12.0 & 20.0 & 50.0 & 18.0 & $-8.0$ & 38.0 & 30.0 & 0.3242 \\
30--40\%  &  0.0 &  0.0 & 54.0 & 46.0 &  0.0 & 46.0 & 46.0 & 0.0182 \\
40--50\%  &  0.0 &  0.0 & 26.0 & 74.0 &  0.0 & 74.0 & 74.0 & 0.0000 \\
50--60\%  &  0.0 &  0.0 & 22.0 & 78.0 &  0.0 & 78.0 & 78.0 & 0.0000 \\
60--70\%  &  0.0 &  0.0 & 26.0 & 74.0 &  0.0 & 74.0 & 74.0 & 0.0000 \\
70--80\%  &  0.0 &  0.0 & 26.0 & 74.0 &  0.0 & 74.0 & 74.0 & 0.0000 \\
80--90\%  &  0.0 &  0.0 & 34.0 & 66.0 &  0.0 & 66.0 & 66.0 & 0.0000 \\
90--100\% &  0.0 &  0.0 & 32.0 & 68.0 &  0.0 & 68.0 & 68.0 & 0.0000 \\
\bottomrule
\end{tabular}
\end{table}

\begin{table}[t]
\caption{Direction of retrieval-induced belief shifts on Llama-3.1-8B, using the same protocol and notation as Table~\ref{tab:belief-direction-qwen}.}
\label{tab:belief-direction-llama}
\centering
\setlength{\tabcolsep}{3.2pt}
\renewcommand{\arraystretch}{1.06}
\begin{tabular}{lrrrrrrrr}
\toprule
JS rank
& Fix
& Break
& W$\rightarrow$W
& R$\rightarrow$R
& Fix$-$Break
& Initial Acc.
& Teacher Acc.
& Mean JS \\
\midrule
0--10\%   & 26.0 & 34.0 & 40.0 &  0.0 & $-8.0$ & 34.0 & 26.0 & 0.6931 \\
10--20\%  & 26.0 & 20.0 & 54.0 &  0.0 &  6.0 & 20.0 & 26.0 & 0.6414 \\
20--30\%  & 26.0 & 16.0 & 56.0 &  2.0 & 10.0 & 18.0 & 28.0 & 0.4223 \\
30--40\%  &  4.0 &  4.0 & 68.0 & 24.0 &  0.0 & 28.0 & 28.0 & 0.1815 \\
40--50\%  &  0.0 &  0.0 & 62.0 & 38.0 &  0.0 & 38.0 & 38.0 & 0.0461 \\
50--60\%  &  0.0 &  0.0 & 38.0 & 62.0 &  0.0 & 62.0 & 62.0 & 0.0000 \\
60--70\%  &  0.0 &  0.0 & 28.0 & 72.0 &  0.0 & 72.0 & 72.0 & 0.0000 \\
70--80\%  &  0.0 &  0.0 & 26.0 & 74.0 &  0.0 & 74.0 & 74.0 & 0.0000 \\
80--90\%  &  0.0 &  0.0 & 18.0 & 82.0 &  0.0 & 82.0 & 82.0 & 0.0000 \\
90--100\% &  0.0 &  0.0 & 18.0 & 82.0 &  0.0 & 82.0 & 82.0 & 0.0000 \\
\bottomrule
\end{tabular}
\end{table}

The comparison separates the magnitude of an information intervention from its direction. In the top 20\% selected by \method{}, Qwen3-4B obtains 36\% Fix and 23\% Break, yielding a positive Fix$-$Break balance of 13 percentage points and increasing teacher accuracy from 23\% to 36\%. The same subset on Llama-3.1-8B contains 26\% Fix and 27\% Break: its Fix$-$Break balance is $-1$ point, and teacher accuracy decreases from 27\% to 26\%. Moreover, 47\% of the selected Llama targets remain wrong before and after retrieval, compared with 41\% for Qwen3. These results show that belief shift successfully concentrates retrieval-sensitive examples on both models, but the induced changes are substantially less corrective on Llama. The negative direction of its highest-ranked interval is particularly important: despite a mean JS close to the maximum $\log 2$, Break exceeds Fix by eight points. Thus, a large realized information flow does not necessarily provide a better pseudo-label when the base model and its retrieved sources share systematic errors.

\subsection{Retrieval-Depth Ablation}
\label{app:retrieval-depth}

\begin{figure}[t]
    \centering
    \includegraphics[width=0.78\linewidth]{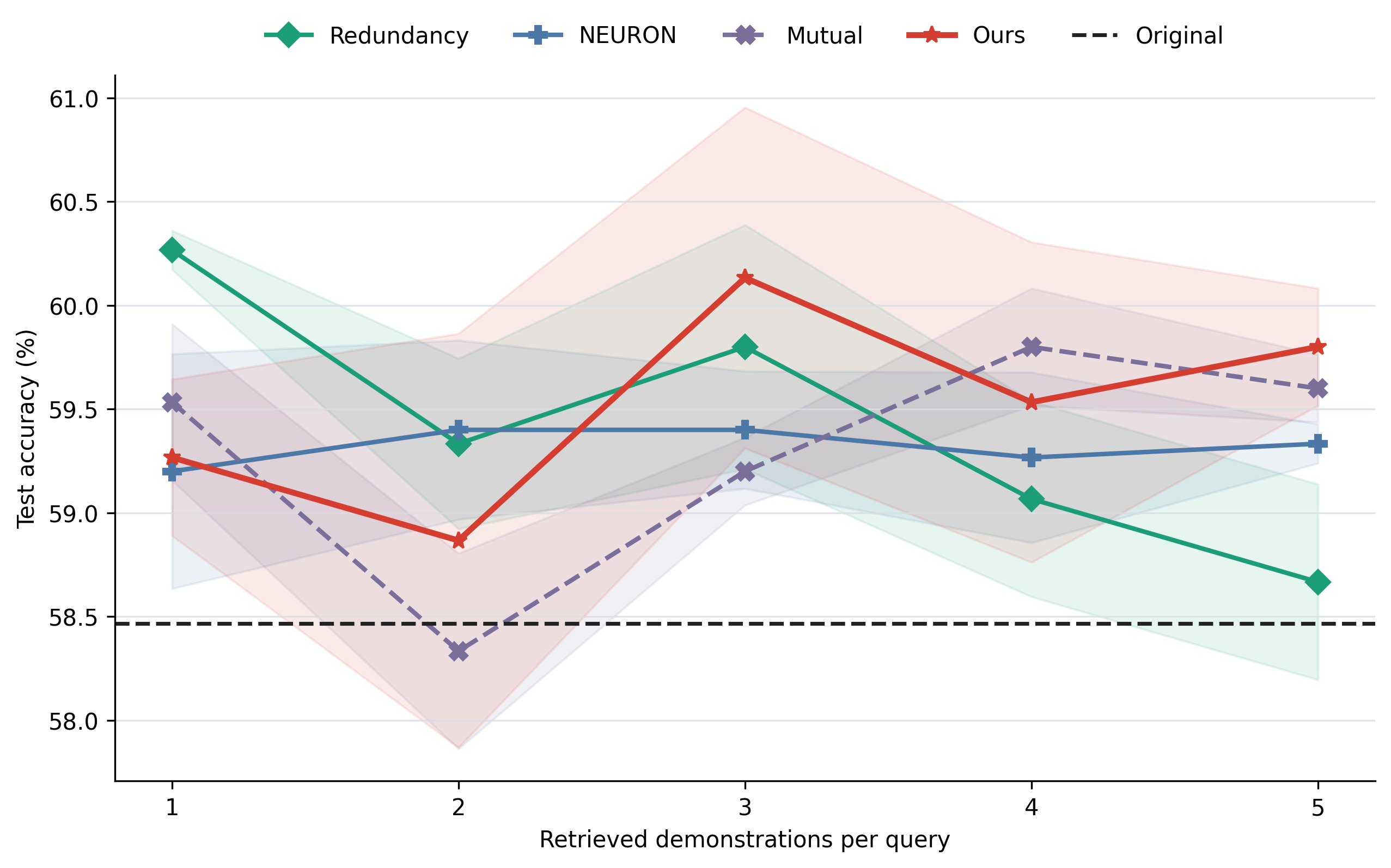}
    \caption{Qwen3-4B-Instruct ablation over the number of retrieved demonstrations per target. Other generation and optimization budgets are fixed.}
    \label{fig:retrieval-k}
\end{figure}

The number of demonstrations trades off potential evidence against context interference. Figure~\ref{fig:retrieval-k} varies $k$ for Qwen3-4B while fixing the selection ratio and optimizer budget. Increasing $k$ initially expands the available source information, but additional sources can be redundant, correlated, or incompatible with the target. The curve therefore measures where the conditional-independence approximation ceases to translate into realized gains.

\subsection{Knowledge Graph Visualization}
\label{app:retrieval-graphs}

We visualize how different retrieval criteria organize the information sources supplied to Qwen3-4B-Instruct. We fix the same target queries across all methods and retrieve three contexts for each target from a shared candidate pool. All questions are projected into a common two-dimensional PCA space using their answer representations. Colored points denote the three knowledge domains, outlined circles mark the fixed target queries, squares mark the retrieved contexts, and edges indicate query--context retrieval relations. Because the projection and target queries are shared, differences between panels arise solely from the retrieval rule.

\begin{figure}[t]
    \centering
    \includegraphics[width=\linewidth]{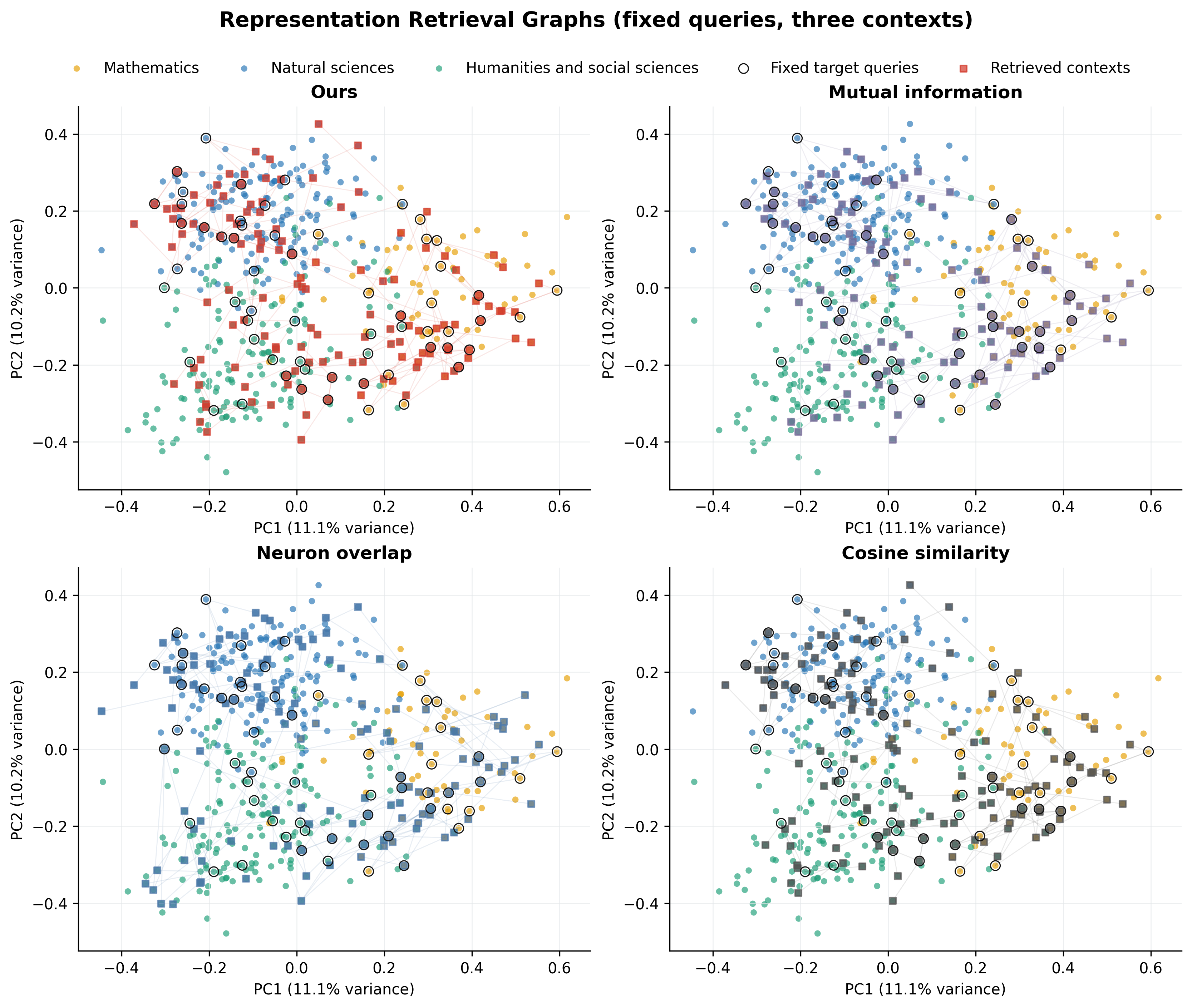}
    \caption{Knowledge-retrieval graphs on Qwen3-4B-Instruct. Each panel uses the same target queries, candidate pool, and retrieval depth $k=3$. Background colors indicate mathematics, natural sciences, and humanities and social sciences. Outlined circles are target queries, squares are retrieved contexts, and connecting edges show the resulting retrieval relations. PC1 and PC2 explain $11.1\%$ and $10.2\%$ of the representation variance, respectively.}
    \label{fig:retrieval-graphs}
\end{figure}

Figure~\ref{fig:retrieval-graphs} reveals distinct retrieval structures. Final-layer cosine similarity primarily follows local neighborhoods in the projected representation space. Greedy mutual-information retrieval introduces broader connections by favoring sources that provide additional information beyond those already selected, while neuron overlap produces relations governed by shared contributing-neuron sets rather than geometric proximity. \method{} exhibits a different balance: it preserves many locally compatible, domain-structured relations but also retrieves sources beyond the nearest visible neighbors when their layerwise decision trajectories align with the query and their source beliefs are sufficiently certain. Consequently, its graph is neither a nearest-neighbor graph nor a maximally dispersed one. It represents query-specific information channels selected jointly by internal computational alignment and source confidence, consistent with the potential-information stage of \method{}.

\section{Baseline Instantiations under the Unified Framework}
\label{app:baseline-details}

The compared methods were originally developed under different supervision and optimization settings. We retain their central selection principles but place them in the same teacher-generation and on-policy distillation framework. Table~\ref{tab:baseline-instantiations} specifies how each method constructs the retrieval context and selects teacher data. Except for \textsc{Original} and \textsc{Full}, every method retains the same fraction $\alpha$ of the candidate pool. All trainable methods use the same number of teacher samples, decoding configuration, optimizer updates, and loss in Equation~\ref{eq:opsd}.

\begin{table}[t]
\caption{Baseline instantiations in the unified comparison framework. Here $k$ is the retrieval depth, $M$ is the number of teacher generations, and $\alpha$ is the selection ratio.}
\label{tab:baseline-instantiations}
\centering
\small
\setlength{\tabcolsep}{4pt}
\renewcommand{\arraystretch}{1.15}
\begin{tabular}{@{}p{0.15\linewidth}p{0.36\linewidth}p{0.41\linewidth}@{}}
\toprule
Method & Retrieval rule & Teacher-data selection rule \\
\midrule
\textsc{Original}
& None.
& No adaptation. \\

\textsc{Full}
& Retrieve the $k$ sources with the largest cosine similarity between final-layer answer representations.
& Retain all generated teacher data. \\

\textsc{Redundancy}
& Use the same final-layer cosine retrieval as \textsc{Full}.
& Iteratively remove the example most similar to another retained teacher representation until $\alpha N$ examples remain. \\

\textsc{Consistency}
& No retrieved demonstrations; instead, we obtain more confident training data through multiple sampling, consistency checks, and self-feedback.

& Retain the $\alpha N$ targets with the largest modal-answer agreement among $M$ generations. \\

\textsc{NEURON}
& Retrieve the $k$ sources with the largest Jaccard overlap between sets of positively contributing neurons.
& Retain the $\alpha N$ targets with the smallest mean number of activated contributing neurons across teacher generations. \\

\textsc{Mutual}
& Retrieve the $k$ sources that provide more conditional mutual information for the target data using a greedy algorithm.
& Retain the $\alpha N$ targets with the largest aggregated retrieval-information score. \\
\bottomrule
\end{tabular}
\end{table}

\paragraph{Redundancy.}
Following the diversity component of DEITA~\citep{liu2024deita}, we remove examples that duplicate information already represented in the retained set. Let $\vz_i$ be the normalized final-layer representation of the selected teacher response for target $i$. For a current retained set $S$, the redundancy of example $i$ is
\begin{equation}
r_i(S)
=
\max_{j\in S\setminus\{i\}}
\cos(\vz_i,\vz_j).
\label{eq:baseline-redundancy}
\end{equation}
Starting from the complete teacher-data pool, the algorithm repeatedly applies
\begin{equation}
i^\star=\arg\max_{i\in S}r_i(S),
\qquad
S\leftarrow S\setminus\{i^\star\},
\label{eq:baseline-redundancy-removal}
\end{equation}
until $|S|=\lceil\alpha N\rceil$. This instantiation isolates representation diversity because annotation-free evaluation provides no external quality score.

\paragraph{NEURON.}
For each response, let $\mathcal{N}_i$ contain the positively contributing neurons retained by token-level and global Top-$K$ attribution. Retrieval follows the activation-overlap principle of NEURON~\citep{chen2026neuron}:
\begin{equation}
J_{ij}
=
\frac{|\mathcal{N}_i\cap\mathcal{N}_j|}
     {|\mathcal{N}_i\cup\mathcal{N}_j|},
\qquad
S_i=\operatorname{TopK}_{j\neq i}(J_{ij},k).
\label{eq:baseline-neuron-retrieval}
\end{equation}
After generating $M$ teacher responses, the selection score is their mean activation-set size,
\begin{equation}
A_i
=
\frac{1}{M}\sum_{m=1}^{M}
\left|\mathcal{N}_i^{(m)}\right|.
\label{eq:baseline-neuron-selection}
\end{equation}
The lowest-scoring $\alpha N$ targets are retained, following the hypothesis that concentrated neuron support indicates a more stable teaching signal.

\paragraph{Mutual.}
The Mutual baseline adapts conditional mutual-information selection
from chain-of-thought distillation~\citep{chen2024mutual}. Let
\begin{equation}
\mathbf{z}_i
=
\operatorname{vec}\!\left(
\mathbf{h}_i^{(1)},\ldots,\mathbf{h}_i^{(L)}
\right)
\label{eq:mutual-representation}
\end{equation}
denote the representation of target $i$, obtained by concatenating its
normalized layerwise answer representations. Rather than ranking
sources independently, Mutual constructs the retrieved set greedily.
Starting from $S_i^{(0)}=\varnothing$, its $t$-th retrieval step is
\begin{equation}
j_t^\star
=
\arg\max_{
j\notin S_i^{(t-1)}\cup\{i\}
}
\widehat{I}\!\left(
\mathbf{z}_i;
\mathbf{z}_j
\mid
\mathbf{z}_{S_i^{(t-1)}}
\right),
\qquad
S_i^{(t)}
=
S_i^{(t-1)}\cup\{j_t^\star\},
\label{eq:mutual-greedy-retrieval}
\end{equation}
for $t=1,\ldots,k$. Here,
$\widehat{I}(\mathbf{z}_i;\mathbf{z}_j
\mid\mathbf{z}_{S})$ estimates the additional representation-level
information that candidate $j$ provides about target $i$ after
conditioning on the sources already included in $S$. The first step
reduces to ordinary mutual-information maximization because
$S_i^{(0)}$ is empty.

The information assigned to the resulting context is the cumulative
marginal gain
\begin{equation}
U_i^{\mathrm{mut}}
=
\sum_{t=1}^{k}
\widehat{I}\!\left(
\mathbf{z}_i;
\mathbf{z}_{j_t^\star}
\mid
\mathbf{z}_{S_i^{(t-1)}}
\right).
\label{eq:mutual-selection}
\end{equation}
Equivalently, when the estimator is consistent with the chain rule,
$U_i^{\mathrm{mut}}$ estimates
$\widehat{I}(\mathbf{z}_i;\mathbf{z}_{S_i^{(k)}})$.
The retrieved context is $S_i=S_i^{(k)}$, and the $\alpha N$ targets
with the largest $U_i^{\mathrm{mut}}$ are retained for distillation.

Mutual therefore accounts for redundancy through conditional,
set-dependent retrieval, but both retrieval and target selection are
completed before observing the retrieval-conditioned teacher answers.
In contrast, \method{} uses a pre-generation score to construct the
context and then selects targets according to the semantic belief
change realized after that context is applied.

\section{Reproducibility Details}
\subsection{Implementation}
\label{app:implementation}

All experiments use the same unlabeled training pool and a disjoint 500-question test set. The source ordering and split construction are fixed across methods, and gold answers are accessed only during evaluation and the post-hoc direction analysis. Unless otherwise stated, candidate preparation uses one initial generation per question with a maximum prompt length of 2,048 tokens. Retrieval-conditioned teacher generation uses a maximum prompt length of 4,096 tokens and produces five responses per target. Both stages use a maximum of 256 generated tokens, temperature $0.6$, and nucleus sampling with $p=0.95$. The layerwise trajectory representation is projected to 64 dimensions with a fixed Rademacher projection generated from seed 42. The default retrieval depth is three, and the selector retains the top 20\% of the question pool. Candidate generation, retrieval, and selection are completed before optimization and are held fixed within each training run.

We optimize rank-16 LoRA adapters with scaling parameter 32 and zero LoRA dropout. For Qwen3-4B, Ministral-3-8B, and Llama-3.1-8B, LoRA is applied to the query, key, value, output, gate, up, and down projections. Qwen3.5 additionally adapts its Gated DeltaNet input and output projections. Training uses AdamW with learning rate $2.5\times10^{-6}$, zero weight decay, cosine decay, a 3\% warm-up ratio, gradient-norm clipping at 1.0, and minibatches of four on-policy trajectories. Each trajectory is limited to 2048 generated tokens, while the combined teacher sequence is truncated to 8,192 tokens. The token-level reverse KL is averaged over response positions and then over the minibatch. The EMA teacher is initialized from the student and updated after every optimizer step as
$\bar{\theta}\leftarrow(1-\gamma)\bar{\theta}+\gamma\theta$,
where $\gamma=0.00501886$, corresponding to an EMA retention coefficient of approximately $0.995$.

The complete update curves use 2,000 on-policy trajectories, corresponding to 500 optimizer updates, and are evaluated every 50 updates. Table~\ref{tab:main} reports the checkpoint at 250 updates. Evaluation uses three independently sampled responses per test question with temperature $0.6$, top-$p=0.95$, and at most 2048 generated tokens. Avg@3 averages correctness over all three rollout columns, its reported standard deviation is computed across the three rollout-level accuracies, and Maj@3 evaluates their majority answer. All model computation uses bfloat16 on one NVIDIA Virtual GPU (vGPU) 48GB. The random seed is fixed to 42 for candidate preparation, subset ordering, optimization, and evaluation; method-specific deterministic offsets are used for teacher generation so that separate methods do not reuse identical sampling streams. The same candidate pool, generation budget, update budget, decoding parameters, and evaluation set are used by all compared methods.
\subsection{Prompt Templates}
\label{app:prompts}

The structured initial-response prompt used to obtain the decision token and its chain-of-thought is:
\begin{verbatim}
{question}

{labeled_options}

Solve the problem internally, then return exactly one valid JSON
object and no other text. Use this schema:
{"knowledge":["keyword 1","keyword 2"],
 "method":"one concise sentence describing the solution method",
 "result":"the complete text of the selected option",
 "answer":"X"}.
Knowledge must contain only a few short keywords, method must be
exactly one concise sentence, result must reproduce the selected
option content, and answer must be its option letter.
\end{verbatim}

\section{Case Studies}
\label{app:examples}

We report 3 Qwen3 examples from SciKnowEval and 5 from MMLU-Pro. Each entry shows the question, retrieved demonstrations, the no-retrieval response and normalized semantic belief, the retrieval-conditioned response and belief, and the resulting belief shift. Through the case study, we can see how the metrics in InFlow are computed and, in a concrete example, examine both the effectiveness and the limitations of the information-gap selection signal in correcting erroneous student responses without access to ground-truth labels.

\subsection{Natural Sciences}
\begin{tcolorbox}[
    breakable,
    colback=gray!5,
    colframe=gray!40,
    boxrule=0.5pt,
    arc=1mm,
    left=5pt,
    right=5pt,
    top=5pt,
    bottom=5pt,
    before skip=6pt,
    after skip=6pt,
    fontupper=\scriptsize
]

\textbf{Case Study 1: sciknoweval:Biology:00000006}

\textbf{Question:}
In the context of MHC binding and peptide predictions, what term refers to the subset of peptide residues that interact with the binding groove of MHC?

\textbf{Choices:}
A: Allele conservation;
B: Binding core;
C: Aggregate adjacency;
D: Canonical adjacency.
\quad
\textbf{Gold:} B

\medskip
\textbf{Initial Response:}

\texttt{\{"knowledge":["MHC binding","peptide interaction","binding groove"],"method":"The term 'binding core' refers to the specific residues of a peptide that directly interact with the MHC binding groove, as defined in peptide-MHC binding studies.","result":"Binding core","answer":"B"\}}

Parsed answer: B;
$p(B)=1.0$;
choice entropy $=1.11\times10^{-13}$;
normalized choice entropy $=8.02\times10^{-14}$;
mean token entropy $=0.144$.

\medskip
\textbf{Retrieved Contexts:}

1. \textit{In the context of bacterial genetics, what term refers to a specific site for transposon Tn7 integration?}
Answer: C;
score $=0.5161$;
certainty $\approx 1.0$.

2. \textit{In the context of HIV-1 research, which of the following is a key determinant for viral entry and cellular tropism?}
Answer: A;
score $=0.5507$;
certainty $\approx 1.0$.

3. \textit{In the context of ion translocation through nanopores, which factor is NOT generally associated with affecting the free energy of ion translocation?}
Answer: A;
score $=0.6569$;
certainty $\approx 1.0$.

\medskip
\textbf{Teacher Generation:}

\textbf{Sample 0:}
\texttt{\{"knowledge":["MHC binding","peptide residues","binding groove","binding core"],"method":"Determined the correct term describing the specific peptide residues that interact with the MHC binding groove from standard immunology and peptide binding terminology.","result":"Binding core","answer":"B"\}}

\textbf{Sample 1:}
\texttt{\{"knowledge":["MHC binding","peptide residues","binding groove","binding core"],"method":"Identified the term that specifically describes the peptide residues directly interacting with the MHC binding groove from standard immunology terminology.","result":"Binding core","answer":"B"\}}

\textbf{Sample 2:}
\texttt{\{"knowledge":["MHC binding","peptide residues","binding groove","binding core"],"method":"Determined the correct term describing the specific peptide residues that interact with the MHC binding groove from standard immunology and peptide binding terminology.","result":"Binding core","answer":"B"\}}

\textbf{Sample 3:}
\texttt{\{"knowledge":["MHC binding","peptide residues","binding groove","binding core"],"method":"Identified the correct term describing the specific peptide residues that directly interact with the MHC binding groove from standard immunology and peptide binding terminology.","result":"Binding core","answer":"B"\}}

\textbf{Sample 4:}
\texttt{\{"knowledge":["MHC binding","peptide residues","binding groove","binding core"],"method":"Determined the correct term describing the specific peptide residues that interact with the MHC binding groove from standard immunology and peptide prediction terminology.","result":"Binding core","answer":"B"\}}

Hard vote counts: \{B: 5\};
modal answer: B;
vote agreement: 1.0;
best sample index: 4.

\medskip
\textbf{Data Selection:}
distribution shift JS $=5.55\times10^{-17}$;
selected = false.

\end{tcolorbox}

\begin{tcolorbox}[
    breakable,
    colback=gray!5,
    colframe=gray!40,
    boxrule=0.5pt,
    arc=1mm,
    left=5pt,
    right=5pt,
    top=5pt,
    bottom=5pt,
    before skip=6pt,
    after skip=6pt,
    fontupper=\scriptsize
]

\textbf{Case Study 2: sciknoweval:Chemistry:00008081}

\textbf{Question:}
In the context of microbial thermoelectric systems, what is the primary function of a ThermoElectric Heat Exchanger (TEHE)?

\textbf{Choices:}
A: Microbial metabolic heat conversion;
B: Temperature control and fresh medium input;
C: Cellular density stabilization;
D: Electricity storage and distribution.
\quad
\textbf{Gold:} B

\medskip
\textbf{Initial Response:}

\texttt{\{"knowledge":["heat exchange","temperature regulation","microbial systems"],"method":"The primary function of a TEHE in microbial thermoelectric systems is to facilitate efficient heat transfer between microbial processes and the environment to maintain optimal operating temperatures.","result":"Temperature control and fresh medium input","answer":"B"\}}

Parsed answer: B;
$p(B)=1.0$.

\medskip
\textbf{Retrieved Contexts:}

1. \textit{In the context of cell culture duration and responses in PB design experiments, which term represents the maximum amount of integral viable cell counts (IVCCs) during cell culture?}

Demo answer: C;
gold: C;
score $=0.6495$.

2. \textit{Which of the following behaviors is NOT typically associated with amoeboid swimmers in a fluid environment?}

Demo answer: A;
gold: B;
score $=0.6729$.

3. \textit{In the context of microfluidic devices, what term refers to the ability of a material to absorb and release small molecules like drugs and growth factors when in contact with cell culture media?}

Demo answer: B;
gold: B;
score $=0.8598$.

\medskip
\textbf{Teacher Generation:}

\textbf{Sample 0:}

\texttt{\{"knowledge":["thermal exchange","heat conversion","microbial metabolism"],"method":"The primary function of a ThermoElectric Heat Exchanger in microbial thermoelectric systems is to facilitate the transfer of metabolic heat generated by microbes into usable electrical energy through thermoelectric effects.","result":"A. Microbial metabolic heat conversion","answer":"A"\}}

Parsed answer: A.

\medskip
\textbf{Sample 1:}

\texttt{\{"knowledge":["thermoelectric","heat exchanger","heat conversion","microbial metabolism"],"method":"In microbial thermoelectric systems, the TEHE primarily facilitates the transfer of heat generated by microbial metabolism to generate electrical energy through thermoelectric effects.","result":"A. Microbial metabolic heat conversion","answer":"A"\}}

Parsed answer: A.

\medskip
\textbf{Sample 2:}

\texttt{\{"knowledge":["thermal exchange","heat conversion","microbial metabolism"],"method":"In microbial thermoelectric systems, the TEHE's primary function is to facilitate the transfer of heat generated by microbial metabolism into usable electrical energy through thermal gradients.","result":"A. Microbial metabolic heat conversion","answer":"A"\}}

Parsed answer: A.

\medskip
\textbf{Sample 3:}

\texttt{\{"knowledge":["thermal exchange","heat conversion","microbial metabolism"],"method":"In microbial thermoelectric systems, the ThermoElectric Heat Exchanger (TEHE) facilitates the transfer of heat from microbial metabolic processes to generate electricity, making heat conversion its primary function.","result":"A. Microbial metabolic heat conversion","answer":"A"\}}

Parsed answer: A.

\medskip
\textbf{Sample 4:}

\texttt{\{"knowledge":["thermoelectric","heat exchange","microbial metabolism","temperature control"],"method":"The primary function of a ThermoElectric Heat Exchanger in microbial thermoelectric systems is to facilitate temperature control by exchanging heat generated from microbial metabolism with the environment.","result":"B. Temperature control and fresh medium input","answer":"B"\}}

Parsed answer: B.

\medskip
Hard vote counts: \{A: 4, B: 1\};
modal answer: A;
vote agreement: 0.8;
best sample index: 3.

\medskip
\textbf{Data Selection:}

distribution shift JS $=0.422810$;
selected = false.

\end{tcolorbox}

\begin{tcolorbox}[
    breakable,
    colback=gray!5,
    colframe=gray!40,
    boxrule=0.5pt,
    arc=1mm,
    left=5pt,
    right=5pt,
    top=5pt,
    bottom=5pt,
    before skip=6pt,
    after skip=6pt,
    fontupper=\scriptsize
]

\textbf{Case Study 3: sciknoweval:Physics:00023453}

\textbf{Question:}
Which material has been identified as a prime candidate for use in optical temperature sensors due to its temperature-susceptible refractive index?

\textbf{Choices:}
A: Silicon Carbide;
B: Gallium Nitride;
C: SU-8;
D: Indium Phosphide.
\quad
\textbf{Gold:} C

\medskip
\textbf{Initial Response:}

\texttt{\{"knowledge":["refractive index","temperature sensitivity","optical sensors"],"method":"The material with a temperature-susceptible refractive index is identified by evaluating known optical materials for their sensitivity to temperature changes through refractive index variation.","result":"Indium Phosphide","answer":"D"\}}

Parsed answer: D;
$p(D)=1.0$;
choice entropy $=1.61\times10^{-13}$;
normalized choice entropy $=1.16\times10^{-13}$;
mean token entropy $=0.166$.

\medskip
\textbf{Retrieved Contexts:}

1. \textit{In the context of confocal Brillouin microscopy, what does the term 'Brillouin elastic contrast' primarily indicate?}
Demo answer: A;
gold: A;
score $=0.5813$;
certainty $\approx 1.0$.

2. \textit{A key factor influencing the shift of the emission peak, denoted as $\Delta\omega_2$, in a temperature-dependent wavelength measurement is:}
Demo answer: B;
gold: A;
score $=0.6943$;
certainty $\approx 1.0$.

3. \textit{Which material is often used in nanophotonic applications due to its high refractive index, well-established fabrication methodology, and compatibility with microfabrication techniques?}
Demo answer: C;
gold: A;
score $=0.7064$;
certainty $\approx 1.0$.

\medskip
\textbf{Teacher Generation:}

\textbf{Sample 0:}

\texttt{\{"knowledge":["refractive index","temperature susceptibility","optical temperature sensor"],"method":"Materials with a strong temperature-dependent refractive index are ideal for optical temperature sensing, and SU-8 exhibits such sensitivity due to its thermally responsive optical properties.","result":"SU-8","answer":"C"\}}

Parsed answer: C;
choice entropy $=1.30\times10^{-12}$;
normalized choice entropy $=9.34\times10^{-13}$;
mean token entropy $=0.153$.

\medskip
\textbf{Sample 1:}

\texttt{\{"knowledge":["refractive index","temperature susceptibility","optical temperature sensor"],"method":"Materials with a strong temperature-dependent refractive index are ideal for optical temperature sensing, and SU-8 exhibits such sensitivity due to its thermal and optical properties.","result":"SU-8","answer":"C"\}}

Parsed answer: C;
choice entropy $=1.14\times10^{-12}$;
normalized choice entropy $=8.24\times10^{-13}$;
mean token entropy $=0.170$.

\medskip
\textbf{Sample 2:}

\texttt{\{"knowledge":["refractive index","temperature susceptibility","optical temperature sensor"],"method":"Materials with temperature-sensitive refractive index are preferred for optical temperature sensing, and SU-8 exhibits strong dependence of refractive index on temperature.","result":"SU-8","answer":"C"\}}

Parsed answer: C;
choice entropy $=1.44\times10^{-12}$;
normalized choice entropy $=1.04\times10^{-12}$;
mean token entropy $=0.171$.

\medskip
\textbf{Sample 3:}

\texttt{\{"knowledge":["refractive index","temperature susceptibility","optical temperature sensor"],"method":"Materials with temperature-sensitive refractive index are ideal for optical temperature sensing, and SU-8 exhibits strong dependence of its refractive index on temperature.","result":"SU-8","answer":"C"\}}

Parsed answer: C;
choice entropy $=1.44\times10^{-12}$;
normalized choice entropy $=1.04\times10^{-12}$;
mean token entropy $=0.158$.

\medskip
\textbf{Sample 4:}

\texttt{\{"knowledge":["refractive index","temperature susceptibility","optical temperature sensor"],"method":"Materials with a strong temperature-dependent refractive index are ideal for optical temperature sensing, and SU-8 exhibits such sensitivity due to its thermal and optical properties.","result":"SU-8","answer":"C"\}}

Parsed answer: C;
choice entropy $=1.14\times10^{-12}$;
normalized choice entropy $=8.24\times10^{-13}$;
mean token entropy $=0.170$.

\medskip
Hard vote counts: \{C: 5\};
modal answer: C;
vote agreement: 1.0;
best sample index: 1.

\medskip
\textbf{Data Selection:}

distribution shift JS $=0.693147$;
selected = true;
selection rank = 83;
selection score $=0.693147$.

\end{tcolorbox}

\subsection{Mathematics}
\begin{tcolorbox}[
    breakable,
    colback=gray!5,
    colframe=gray!40,
    boxrule=0.5pt,
    arc=1mm,
    left=5pt,
    right=5pt,
    top=5pt,
    bottom=5pt,
    before skip=6pt,
    after skip=6pt,
    fontupper=\scriptsize
]

\textbf{Case Study 4: mmlu\_pro:Math:00007694}

\textbf{Question:}
Statement 1: There exists a free abelian group of every positive integer rank.
Statement 2: A finitely generated abelian group is free abelian if its Betti number equals the number of elements in some generating set.

\textbf{Choices:}
A: False, False;
B: True, Cannot be determined;
C: False, True;
D: Cannot be determined, True;
E: False, Cannot be determined;
F: Statement 1 is false for certain positive integers, Statement 2 is false;
G: True, False;
H: Statement 1 is true only for certain positive integers, Statement 2 is true;
I: Cannot be determined, False;
J: True, True.
\quad
\textbf{Gold:} J

\medskip
\textbf{Initial Response:}

\texttt{\{"knowledge":["free abelian group","Betti number","finitely generated","rank","generating set"],"method":"A free abelian group of rank n exists for every positive integer n, and the Betti number of a finitely generated abelian group equals the rank of its free part, which must match the number of generators in a minimal generating set for it to be free abelian.","result":"J. True, True","answer":"J"\}}

Parsed answer: J;
$p(J)=1.0$;
choice entropy $=8.04\times10^{-15}$;
normalized choice entropy $=3.49\times10^{-15}$;
mean token entropy $=0.118$.

\medskip
\textbf{Retrieved Contexts:}

1. \textit{The external direct product of cyclic groups is cyclic; the external direct product of $D_3$ and $D_4$ is isomorphic to $D_{12}$.}
Demo answer: C;
gold: I;
score $=0.4948$;
certainty $\approx1.0$.

2. \textit{$\mathbb{R}$ is a splitting field of some polynomial over $\mathbb{Q}$; there is a field with 60 elements.}
Demo answer: B;
gold: H;
score $=0.7709$;
certainty $\approx1.0$.

3. \textit{Every nonzero free abelian group has an infinite number of bases; every free abelian group of rank at least 2 has an infinite number of bases.}
Demo answer: B;
gold: G;
score $=0.9554$;
certainty $\approx1.0$.

\medskip
\textbf{Teacher Generation:}

\textbf{Sample 0:}
\texttt{\{"knowledge":["free abelian group","Betti number","finitely generated","generating set"],"method":"Free abelian groups exist for every positive integer rank by definition, and a finitely generated abelian group is free abelian if and only if it has no torsion, which is equivalent to its Betti number equaling the rank of the free part, not necessarily the number of elements in a generating set.","result":"G. True, False","answer":"G"\}}

\textbf{Sample 1:}
\texttt{\{"knowledge":["free abelian group","Betti number","finitely generated","rank"],"method":"Free abelian groups exist for every positive integer rank by definition, and a finitely generated abelian group is free abelian if and only if it has no torsion, which is equivalent to its Betti number matching the number of generators in a free basis, not just any generating set.","result":"G. True, False","answer":"G"\}}

\textbf{Sample 2:}
\texttt{\{"knowledge":["free abelian group","Betti number","finitely generated","generating set"],"method":"Free abelian groups exist for every positive integer rank by definition, and a finitely generated abelian group is free abelian if and only if it has no torsion, which is equivalent to its Betti number matching the rank of the free part, not necessarily the number of elements in a generating set.","result":"G. True, False","answer":"G"\}}

\textbf{Sample 3:}
\texttt{\{"knowledge":["free abelian group","Betti number","finitely generated","generating set"],"method":"Free abelian groups exist for every positive integer rank by definition, and a finitely generated abelian group is free abelian if and only if it has no torsion, which is equivalent to the Betti number equaling the rank, not necessarily the number of elements in a generating set.","result":"G. True, False","answer":"G"\}}

\textbf{Sample 4:}
\texttt{\{"knowledge":["free abelian group","Betti number","finitely generated","generating set"],"method":"Free abelian groups exist for every positive integer rank by definition, and the Betti number of a finitely generated abelian group equals the rank of its free part; a finitely generated abelian group is free abelian if and only if it has no torsion, which is equivalent to its Betti number matching the number of generators in the free part, not just any generating set.","result":"G. True, False","answer":"G"\}}

Hard vote counts: \{G: 5\};
modal answer: G;
vote agreement: 1.0;
best sample index: 1.

\medskip
\textbf{Data Selection:}
distribution shift JS $=0.693147$;
selected = true;
selection rank = 5;
selection score $=0.693147$.

\end{tcolorbox}

\begin{tcolorbox}[
    breakable,
    colback=gray!5,
    colframe=gray!40,
    boxrule=0.5pt,
    arc=1mm,
    left=5pt,
    right=5pt,
    top=5pt,
    bottom=5pt,
    before skip=6pt,
    after skip=6pt,
    fontupper=\scriptsize
]

\textbf{Case Study 5: mmlu\_pro:Math:00007732}

\textbf{Question:}
What is the units digit in the standard decimal expansion of the number $7^{25}$?

\textbf{Choices:}
A: 8;
B: 2;
C: 9;
D: 4;
E: 1;
F: 0;
G: 7;
H: 3;
I: 5;
J: 6.
\quad
\textbf{Gold:} G

\medskip
\textbf{Initial Response:}

\texttt{\{"knowledge":["units digit","exponent pattern","modular arithmetic"],"method":"The units digit of powers of 7 follows a cycle of 4: 7, 9, 3, 1, so we find the position of 25 in this cycle using modulo 4.","result":"8","answer":"A"\}}

Parsed answer: A;
$p(A)=1.0$.

\medskip
\textbf{Retrieved Contexts:}

1. \textit{A boy found a bicycle lock for which the combination was unknown. The correct combination is a four-digit number, where each digit is selected from 1 through 8. How many different lock combinations are possible?}

Demo answer: A;
gold: G;
score $=0.4884$.

2. \textit{What is the probability that a randomly selected integer in the set $\{1,2,3,\ldots,100\}$ is divisible by 2 and not divisible by 3?}

Demo answer: D;
gold: D;
score $=0.5432$.

3. \textit{For how many positive integral values of $x\leq100$ is $3^x-x^2$ divisible by 5?}

Demo answer: B;
gold: D;
score $=0.6307$.

\medskip
\textbf{Teacher Generation:}

\textbf{Sample 0:}

\texttt{\{"knowledge":["units digit","pattern cycle","exponent modulo"],"method":"Determine the cycle of the units digit of powers of 7, which repeats every 4 terms: 7, 9, 3, 1, and find the position of 7\^25 in the cycle","result":"1","answer":"E"\}}

Parsed answer: E.

\medskip
\textbf{Sample 1:}

\texttt{\{"knowledge":["units digit","pattern cycle","exponent modulo"],"method":"Determine the cyclic pattern of the units digit of powers of 7, which repeats every 4 terms","result":"8","answer":"A"\}}

Parsed answer: A.

\medskip
\textbf{Sample 2:}

\texttt{\{"knowledge":["units digit","pattern cycle","exponent modulo"],"method":"Determine the cycle of the units digit of powers of 7, which repeats every 4 terms: 7, 9, 3, 1, and find the position of 25 in the cycle","result":"1","answer":"E"\}}

Parsed answer: E.

\medskip
\textbf{Sample 3:}

\texttt{\{"knowledge":["units digit","pattern cycle","exponent modulo"],"method":"Identify the cyclic pattern of the units digit of powers of 7, which repeats every 4 exponents","result":"8","answer":"A"\}}

Parsed answer: A.

\medskip
\textbf{Sample 4:}

\texttt{\{"knowledge":["units digit","pattern cycle","powers of 7"],"method":"Determine the cyclic pattern of the units digit of powers of 7 and find the position of 7\^25 in the cycle","result":"7","answer":"G"\}}

Parsed answer: G.

\medskip
Hard vote counts: \{E: 2, A: 2, G: 1\};
modal answer: A;
vote agreement: 0.4;
best sample index: 1.

\medskip
\textbf{Data Selection:}

distribution shift JS $=0.274358$;
selected = false.

\end{tcolorbox}

\subsection{Humanities \& Social Sciences}

\begin{tcolorbox}[
    breakable,
    colback=gray!5,
    colframe=gray!40,
    boxrule=0.5pt,
    arc=1mm,
    left=5pt,
    right=5pt,
    top=5pt,
    bottom=5pt,
    before skip=6pt,
    after skip=6pt,
    fontupper=\scriptsize
]

\textbf{Case Study 6: mmlu\_pro:Psychology:00002027}

\textbf{Question:}
The part of the neuron that directs synthesis of neurotransmitters and proteins is the

\textbf{Choices:}
A: axon terminal;
B: axon;
C: synaptic cleft;
D: neuroglia;
E: myelin sheath;
F: schwann cell;
G: cell body;
H: axon hillock;
I: nodes of Ranvier;
J: dendrite.
\quad
\textbf{Gold:} G

\medskip
\textbf{Initial Response:}

\texttt{\{"knowledge":["cell body","neurotransmitter","protein synthesis"],"method":"The cell body contains the nucleus and organelles responsible for synthesizing neurotransmitters and proteins.","result":"cell body","answer":"G"\}}

Parsed answer: G;
$p(G)=1.0$;
choice entropy $=1.71\times10^{-14}$;
normalized choice entropy $=7.42\times10^{-15}$;
mean token entropy $=0.0598$.

\medskip
\textbf{Retrieved Contexts:}

1. \textit{In the context of ophthalmic imaging, what is the primary purpose of utilizing targeted spectroscopy?}
Demo answer: D;
gold: D;
score $=0.4546$;
certainty $\approx1.0$.

2. \textit{In the context of fungal genetics, what term refers to the process of sexual reproduction between different species?}
Demo answer: B;
gold: B;
score $=0.4578$;
certainty $\approx1.0$.

3. \textit{In the context of cell movement, what primarily drives the protoplasmic flow that deforms the cell body and propels the cell?}
Demo answer: B;
gold: B;
score $=0.4970$;
certainty $\approx1.0$.

\medskip
\textbf{Teacher Generation:}

\textbf{Sample 0:}

\texttt{\{"knowledge":["neurotransmitter synthesis","protein synthesis","cell body"],"method":"The cell body of a neuron contains the organelles and machinery necessary for synthesizing neurotransmitters and proteins.","result":"cell body","answer":"G"\}}

\textbf{Sample 1:}

\texttt{\{"knowledge":["neurotransmitter synthesis","protein synthesis","cell body"],"method":"The cell body of a neuron contains the organelles and machinery responsible for synthesizing neurotransmitters and proteins.","result":"cell body","answer":"G"\}}

\textbf{Sample 2:}

\texttt{\{"knowledge":["neurotransmitter synthesis","protein synthesis","cell body"],"method":"The cell body of a neuron contains the organelles and machinery necessary for synthesizing neurotransmitters and proteins.","result":"cell body","answer":"G"\}}

\textbf{Sample 3:}

\texttt{\{"knowledge":["neurotransmitter synthesis","protein synthesis","cell body"],"method":"The cell body of a neuron contains the organelles and machinery necessary for synthesizing neurotransmitters and proteins.","result":"cell body","answer":"G"\}}

\textbf{Sample 4:}

\texttt{\{"knowledge":["neurotransmitter synthesis","protein synthesis","cell body"],"method":"The cell body of a neuron contains the organelles and machinery necessary for synthesizing neurotransmitters and proteins.","result":"cell body","answer":"G"\}}

\medskip
Hard vote counts: \{G: 5\};
modal answer: G;
vote agreement: 1.0;
best sample index: 0.

\medskip
\textbf{Data Selection:}

distribution shift JS $=5.55\times10^{-17}$;
selected = false.

\end{tcolorbox}

\begin{tcolorbox}[
    breakable,
    colback=gray!5,
    colframe=gray!40,
    boxrule=0.5pt,
    arc=1mm,
    left=5pt,
    right=5pt,
    top=5pt,
    bottom=5pt,
    before skip=6pt,
    after skip=6pt,
    fontupper=\scriptsize
]

\textbf{Case Study 7: mmlu\_pro:Law:00000878}

\textbf{Question:}
An uncle owns his property in fee simple. The uncle executes and delivers a deed to the property with the following granting clause: ``To my nephew and my niece for the life of my nephew, then to my gardener.'' As a result of the grant, which of the following is the most accurate statement regarding the ownership rights of the property?

\textbf{Choices:}
A: The nephew has a life estate pur autre vie, the niece has a vested remainder, and the gardener has a life estate;
B: The nephew and niece are tenants in common for the life of the first of them to die, the nephew and niece have contingent remainders in fee simple defeasible, and the gardener has an executory interest;
C: The nephew and niece are joint tenants with rights of survivorship, and the gardener has a contingent remainder;
D: The nephew and niece are tenants in common, and the gardener has a vested remainder in fee simple;
E: The nephew and niece are joint tenants in fee simple defeasible, and the gardener has an executory interest;
F: The nephew has a vested remainder, the niece has a life estate, and the gardener has a life estate pur autre vie;
G: The nephew and niece both have life estates pur autre vie, and the gardener has a contingent remainder;
H: The nephew has a life estate, the niece has a life estate pur autre vie, and the gardener has a vested remainder;
I: The nephew has a life estate, the niece has a vested remainder, and the gardener has a life estate pur autre vie;
J: The nephew has a life estate, the niece also has a life estate, and the gardener has a vested remainder.
\quad
\textbf{Gold:} H

\medskip
\textbf{Initial Response:}

\texttt{\{"knowledge":["life estate","vested remainder","executory interest","pur autre vie","tenants in common"],"method":"The granting clause specifies a life estate to the nephew, a remainder interest to the niece, and a life estate to the gardener, with the nephew's life determining the termination of the grant, thus creating a life estate pur autre vie for the nephew and a vested remainder for the niece, with the gardener receiving an executory interest upon the nephew's death.","result":"The nephew has a life estate pur autre vie, the niece has a vested remainder, and the gardener has an executory interest.","answer":"B"\}}

Parsed answer: B;
$p(B)=0.9241$;
choice entropy $=0.2685$;
normalized choice entropy $=0.1166$;
mean token entropy $=0.2498$.

\medskip
\textbf{Retrieved Contexts:}

1. \textit{A general contractor invited a carpenter and others to bid on carpentry work and agreed to give the carpenter the job if his bid was lowest and the general contractor received the main contract. Which fact best supports the carpenter's position that the general contractor is obligated to award him the subcontract?}

Demo answer: A;
gold: I;
score $=0.6633$;
certainty $\approx1.0$.

2. \textit{A farmer repeatedly used a road across a widower's land to reach the widower's daughter's property. If the widower prevails in an action to enjoin the farmer's use, it will be because the farmer's use was:}

Demo answer: A;
gold: G;
score $=0.6869$;
certainty $\approx1.0$.

3. \textit{Two cousins held real estate as joint tenants with right of survivorship. One cousin's interest became subject to a judgment lien and he later died. Will the court enforce the lien against the surviving cousin's sale of the property?}

Demo answer: D;
gold: D;
score $=0.7726$;
certainty $\approx1.0$.

\medskip
\textbf{Teacher Generation:}

\textbf{Sample 0:}

\texttt{\{"knowledge":["life estate","vested remainder","executory interest","pur autre vie","tenants in common"],"method":"The grant creates a life estate in the nephew, with a remainder to the niece, and a contingent remainder to the gardener, but the phrase 'for the life of my nephew' indicates a life estate in the nephew, and the remainder to the niece is not a life estate but a vested remainder, with the gardener's interest being an executory interest that only takes effect if the nephew dies and the niece does not inherit.","result":"The nephew has a life estate, the niece has a vested remainder, and the gardener has a life estate pur autre vie.","answer":"I"\}}

Parsed answer: I;
choice entropy $=4.87\times10^{-12}$;
normalized choice entropy $=2.11\times10^{-12}$;
mean token entropy $=0.3032$.

\medskip
\textbf{Sample 1:}

\texttt{\{"knowledge":["life estate","vested remainder","executory interest","pur autre vie","tenants in common"],"method":"The grant creates a life estate to the nephew, a remainder interest to the niece (as a vested remainder), and an executory interest to the gardener, which only takes effect if the nephew dies before the niece, thus requiring interpretation of the phrase 'for the life of my nephew, then to my gardener' as a life estate to the nephew with a contingent remainder to the gardener, but standard interpretation treats 'then to' as an executory interest, not a life estate for the gardener.","result":"The nephew and niece are tenants in common for the life of the first of them to die, the nephew and niece have contingent remainders in fee simple defeasible, and the gardener has an executory interest.","answer":"B"\}}

Parsed answer: B;
choice entropy $=1.49\times10^{-13}$;
normalized choice entropy $=6.48\times10^{-14}$;
mean token entropy $=0.2706$.

\medskip
\textbf{Sample 2:}

\texttt{\{"knowledge":["life estate","vested remainder","executory interest","pur autre vie","tenants in common"],"method":"The grant creates a life estate to the nephew, with the niece receiving a vested remainder, and the gardener receiving an executory interest that takes effect only if the nephew dies; the phrase 'for the life of my nephew' means the nephew holds a life estate, and the niece gets a remainder interest after the nephew's death, with the gardener as a contingent beneficiary only if the nephew dies.","result":"The nephew has a life estate, the niece has a vested remainder, and the gardener has a life estate pur autre vie.","answer":"I"\}}

Parsed answer: I;
choice entropy $=6.53\times10^{-12}$;
normalized choice entropy $=2.84\times10^{-12}$;
mean token entropy $=0.2715$.

\medskip
\textbf{Sample 3:}

\texttt{\{"knowledge":["life estate","vested remainder","executory interest","pur autre vie","tenants in common"],"method":"The grant creates a life estate in the nephew, a vested remainder in the niece, and an executory interest for the gardener, with the nephew's life estate being 'pur autre vie' because it is tied to the life of the nephew, not the first to die.","result":"The nephew has a life estate pur autre vie, the niece has a vested remainder, and the gardener has a life estate.","answer":"I"\}}

Parsed answer: I;
choice entropy $=0.0140$;
normalized choice entropy $=0.00607$;
mean token entropy $=0.2070$.

\medskip
\textbf{Sample 4:}

\texttt{\{"knowledge":["life estate","vested remainder","executory interest","pur autre vie","tenants in common"],"method":"The grant creates a life estate to the nephew, a remainder to the niece, and an executory interest to the gardener, with the nephew's life estate being the primary interest; the phrase 'for the life of my nephew' establishes a life estate, and 'then to my gardener' indicates an executory interest that takes effect after the nephew's death, not a vested remainder for the niece.","result":"The nephew has a life estate, the niece has a vested remainder, and the gardener has a life estate pur autre vie.","answer":"I"\}}

Parsed answer: I;
choice entropy $=1.83\times10^{-12}$;
normalized choice entropy $=7.93\times10^{-13}$;
mean token entropy $=0.2444$.

\medskip
Hard vote counts: \{I: 4, B: 1\};
modal answer: I;
vote agreement: 0.8;
best sample index: 4.

\medskip
\textbf{Data Selection:}

distribution shift JS $=0.428766$;
selected = false.

\end{tcolorbox}

\begin{tcolorbox}[
    breakable,
    colback=gray!5,
    colframe=gray!40,
    boxrule=0.5pt,
    arc=1mm,
    left=5pt,
    right=5pt,
    top=5pt,
    bottom=5pt,
    before skip=6pt,
    after skip=6pt,
    fontupper=\scriptsize
]

\textbf{Case Study 8: mmlu\_pro:Global Facts:00005309}

\textbf{Question:}
In 2016, what percentage of the population of South Sudan had access to electricity?

\textbf{Choices:}
A: 79\%;
B: 49\%;
C: 69\%;
D: 9\%;
E: 59\%;
F: 39\%;
G: 19\%;
H: 29\%;
I: 89\%;
J: 99\%.
\quad
\textbf{Gold:} D

\medskip
\textbf{Initial Response:}

\texttt{\{"knowledge":["South Sudan","2016","electricity access"],"method":"The percentage of the population with access to electricity in South Sudan in 2016 was reported by international sources as 49\% based on data from the World Bank and United Nations Development Programme.","result":"49\%","answer":"B"\}}

Parsed answer: B;
$p(B)=1.0$;
choice entropy $=1.24\times10^{-13}$;
normalized choice entropy $=5.39\times10^{-14}$;
mean token entropy $=0.152$.

\medskip
\textbf{Retrieved Contexts:}

1. \textit{As of 2019, about what share of electricity came from fossil fuels in the United States?}

Demo answer: G;
gold: G;
score $=0.6536$;
certainty $\approx1.0$.

2. \textit{As of 2015, agriculture made up about what percentage of total Indian GDP?}

Demo answer: H;
gold: H;
score $=0.7390$;
certainty $\approx1.0$.

3. \textit{Which of the following countries invested the largest fraction of its GDP in renewable energy in 2015?}

Demo answer: G;
gold: C;
score $=0.7405$;
certainty $\approx1.0$.

\medskip
\textbf{Teacher Generation:}

\textbf{Sample 0:}

\texttt{\{"knowledge":["South Sudan","electricity access","2016"],"method":"According to the International Energy Agency's 2016 data, approximately 19\% of South Sudan's population had access to electricity.","result":"G. 19\%","answer":"G"\}}

Parsed answer: G;
choice entropy $=4.04\times10^{-15}$;
normalized choice entropy $=1.75\times10^{-15}$;
mean token entropy $=0.0679$.

\medskip
\textbf{Sample 1:}

\texttt{\{"knowledge":["South Sudan","electricity access","2016"],"method":"According to the International Energy Agency's data, in 2016, only about 19\% of South Sudan's population had access to electricity.","result":"G. 19\%","answer":"G"\}}

Parsed answer: G;
choice entropy $=4.69\times10^{-15}$;
normalized choice entropy $=2.04\times10^{-15}$;
mean token entropy $=0.0648$.

\medskip
\textbf{Sample 2:}

\texttt{\{"knowledge":["South Sudan","electricity access","2016"],"method":"According to the International Energy Agency, in 2016, only about 19\% of South Sudan's population had access to electricity.","result":"G. 19\%","answer":"G"\}}

Parsed answer: G;
choice entropy $=4.13\times10^{-15}$;
normalized choice entropy $=1.79\times10^{-15}$;
mean token entropy $=0.0562$.

\medskip
\textbf{Sample 3:}

\texttt{\{"knowledge":["South Sudan","electricity access","2016"],"method":"According to the International Energy Agency, in 2016, approximately 19\% of South Sudan's population had access to electricity.","result":"G. 19\%","answer":"G"\}}

Parsed answer: G;
choice entropy $=2.68\times10^{-15}$;
normalized choice entropy $=1.16\times10^{-15}$;
mean token entropy $=0.0568$.

\medskip
\textbf{Sample 4:}

\texttt{\{"knowledge":["South Sudan","electricity access","2016"],"method":"According to the International Energy Agency, in 2016, approximately 19\% of South Sudan's population had access to electricity.","result":"G. 19\%","answer":"G"\}}

Parsed answer: G;
choice entropy $=2.68\times10^{-15}$;
normalized choice entropy $=1.16\times10^{-15}$;
mean token entropy $=0.0568$.

\medskip
Hard vote counts: \{G: 5\};
modal answer: G;
vote agreement: 1.0;
best sample index: 3.

\medskip
\textbf{Data Selection:}

distribution shift JS $=0.693147$;
selected = true;
selection rank = 17;
selection score $=0.693147$.

\end{tcolorbox}

\end{document}